%% file: main.tex
\documentclass[]{kiai}
\usepackage{amsmath,amssymb,mathtools,amsthm}
\usepackage{graphicx}
\usepackage{natbib}
\usepackage[hidelinks,breaklinks]{hyperref}
\usepackage{subfig}
\usepackage{outlines}
\usepackage{nicefrac}
\usepackage{float}
\usepackage{algorithm}
\usepackage{algpseudocode}
\usepackage{enumitem}
\usepackage{microtype}
\usepackage{booktabs}
\usepackage{xcolor}
\usepackage{cancel}
\usepackage[hidelinks]{hyperref}
\usepackage{multirow}

\newtheorem{theorem}{Theorem}[]

\newtheorem{remark1}[theorem]{Remark}
\newtheorem{atheorem}{Theorem}[subsection]
\newtheorem{alemma}[atheorem]{Lemma}
\newtheorem{acorollary}[atheorem]{Corollary}

\newtheorem{aremark1}[atheorem]{Remark}
\newenvironment{aremark}{\begin{aremark1}\rm}{\end{aremark1}}

\theoremstyle{plain}

\theoremstyle{definition}

\theoremstyle{remark}

\newcommand{\R}{\mathbb{R}}

\newcommand{\Prob}{\mathbb P}
\newcommand{\X}{\mathcal X}

\newcommand{\Lfanc}{\mathcal L}
\newcommand{\Zaux}{\mathcal{Z}}

\title{Branching Flows: Discrete, Continuous, and Manifold Flow Matching with Splits and Deletions}
\author[1]{Lukas Billera*}
\author[1]{Hedwig Nora Nordlinder*}
\author[1]{Jack Collier Ryder}
\author[1]{Anton Oresten}
\author[1]{Aron Stålmarck}
\author[1]{Theodor Mosetti Björk}
\author[1]{Ben Murrell}

\affiliation[1]{Department of Microbiology, Tumor and Cell Biology, Karolinska Institutet}
\affiliation[*]{Contributed equally}
\date{November 12, 2025}
\correspondence{\email{benjamin.murrell@ki.se}}

\begin{document}

\abstract{  Diffusion and flow matching approaches to generative modeling have shown promise in domains where the number of elements in a state is fixed in advance (e.g. images), but require \emph{ad hoc} solutions when, for example, the length of a response from a large language model, the number of atoms in a molecule, or the number of amino acids in a protein chain is not known \emph{a priori}. Here we propose Branching Flows, a generative modeling framework that, like diffusion and flow matching approaches, transports a simple distribution to the data distribution. But in Branching Flows, the elements in the state evolve over a forest of binary trees, branching and dying stochastically with rates that are learned by the model. This allows the model to control, during generation, the number of elements in the sequence. We show that Branching Flows can compose with any flow matching base process on discrete sets, continuous Euclidean spaces, Riemannian manifolds, and `multimodal' product spaces that mix these components, and we demonstrate distribution matching on small molecules and antibody sequences, and that this scales to complicated domains such as protein structures.}

\maketitle

\vspace{-.1em}

\section{Introduction}

\begin{figure}[t]
    \centering{\includegraphics[width=1\textwidth]{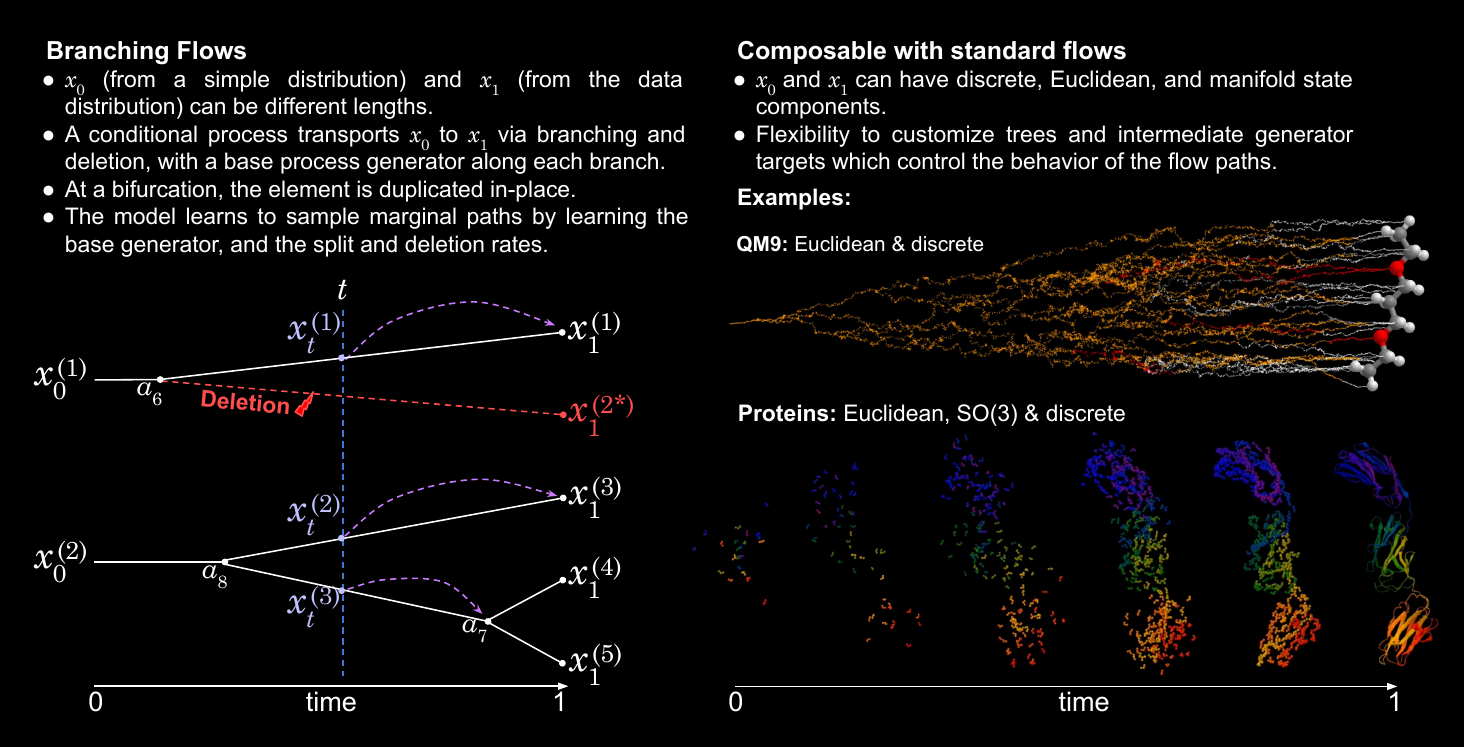}}
    \caption{\textbf{Graphical Abstract.}
    }\label{fig:abstract}
\end{figure}

For many applications that involve generating a sequence of elements, it isn't possible to know the required number of elements in advance. For sequences of discrete tokens this is well addressed by autoregressive (AR) models which control length by sampling stop tokens. Continuous AR models are possible \citep{Billera2024TheCL} but underexplored, and current examples are limited in the range of problems they can address. Diffusion and flow matching methods avoid left-to-right AR structure and excel for structured grids and fixed-length sequences, but sacrifice the ability to naturally handle length variation.

One challenge not naturally addressed by fixed-length diffusion/flow models nor AR models (especially in the continuous state context) is what we call the `unknown-length infix sampling' problem, where a segment of unknown length must be inserted into a sequence, conditioned on \emph{a priori} known flanking regions. As a concrete example from protein design: conditioned on framework regions and some binding pose, if you wish to design the complementarity-determining regions of an antibody you must first specify how long they must be, which is absurd. There is a `soft' version of this problem that arises for fixed-length diffusion/flow models, where flanking regions happen to resolve early during an inference trajectory, and the model cannot control the number of elements spanning them.

In the diffusion and flow matching context, Edit Flows addresses length variation by modeling variable-length discrete sequences via insertions and deletions \citep{Havasi2025EditFF, OneFlowNguyen} and, in the purely-continuous domain, one attempt to develop variable-length diffusion models \citep{Campbell2023TransDimensionalGM} relies on an approximation related to the addition of new elements, and it is unclear how this scales to more complex examples.

Here we present Branching Flows, a framework for generative modeling over variable-length sequences where the elements can be continuous, manifold-valued, discrete, or combinations of these. This is formulated in Generator Matching \citep{Holderrieth2024GeneratorMG}, where a conditional process transports samples from a simple distribution to the data distribution. The generator of this is learned, in expectation, yielding a process whose marginals match those of the conditional process. Branching Flows augments a base Markov generator on an element space with a branching and deletion process. A forest of trees is used to couple a sample from a simple distribution (with elements at the roots) with the data distribution (elements at the leaves). Like the processes used in phylogenetics \citep{felsenstein1981evolutionary}, the elements evolve independently along each branch but duplicate and decouple at bifurcations in the tree. This construction, composed with optional deletions, allows us to train a generative model over sequences of varying length.

\section{Preliminaries}

\subsection{Generator Matching and Auxiliary Generator Matching}

Following Generator Matching \citep{Holderrieth2024GeneratorMG,Lipman2024FlowMG}, for \(t\in[0,1]\), we seek a probability path \(p_t\) that interpolates between two fixed marginals \(p\) and \(q\), where $p$ is a distribution that is easy to sample from, and $q$ is the data distribution. This goes via a conditional-path construction: Conditional on an auxiliary variable \(z\in\mathcal{Z}\) with marginal distribution \(p_Z\) and a family of conditional probability measures \(p_{t\mid Z}(\mathrm{d}x\mid z)\), the corresponding marginal path can be described by hierarchical sampling
\begin{equation*}
     Z\sim p_Z,\quad X_t|(Z=z)\sim p_{t\mid Z}(\mathrm{d}x\mid z) \ \Rightarrow\ X_t\sim p_t(\mathrm{d}x).
\end{equation*}
Choose the conditional family so that the boundary constraints \(p_0=p\) and \(p_1=q\) hold after marginalization. This can be achieved by sampling an $X_0 \sim p$ and an $X_1 \sim q$ and having the process start at $X_0$ and end at $X_1$. The goal is then to parameterize a generator $\Lfanc_t$ by a neural network $\Lfanc_t^\theta$ and train it to generate the marginal path $p_t$. This is done by first drawing samples from the conditional path $X_t\sim p_{t\mid Z}(\mathrm{d}x\mid z)$, and then training $\Lfanc_t^\theta$ against the generators of the conditional paths $\Lfanc_t^z$. Taking gradient steps under a `conditional generator matching' loss (a Bregman divergence) encourages $\Lfanc_t^\theta$ to learn the expectation of the conditional generator which, via a gradient identity, generates the marginal path $p_t$. 

By \emph{`Auxiliary Generator Matching'} we mean: augment our process with \( G_t \) (with realisations denoted $g_t$) for each $t\in [0,1]$, yielding joint conditional paths $(X_t, G_t)|(Z=z) \sim  p_{t\mid Z}(\mathrm{d}x,\mathrm{d}g\mid z)$, but train a model that marginalizes over $G_t$ and $Z$, learning a generator on $X_t$ with marginals matching $X_t\sim p_t(\mathrm{d}x)$. This was used in Edit Flows \citep{Havasi2025EditFF} for the discrete case, but see Appendix~\ref{app:auggm} for our setting. 

\begin{figure*}[h!]
    \centering{\includegraphics[width=0.49\textwidth]{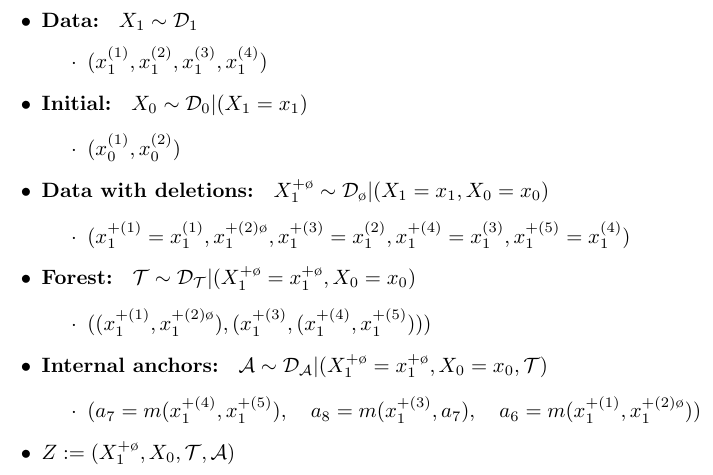}}
    \centering{\raisebox{.00\height}{\includegraphics[width=0.49\textwidth]{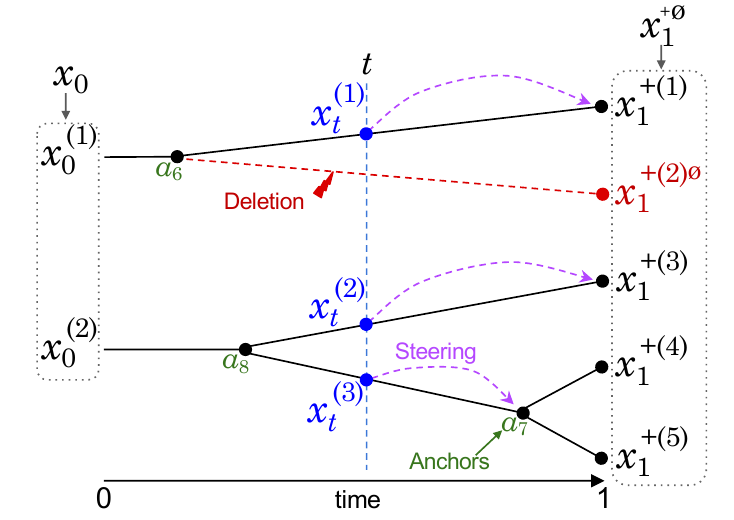}}}
    \caption{\textbf{Branching Flows construction.} Left outlines the sampling of $Z$ when $x_1$ has 4 elements and $x_0$ has two elements, for a conditional bridge that will incur a single deletion (element $x_1^{+(2)ø})$). $m(.,.)$ derives an anchor from two children, which could be e.g. a midpoint for continuous states and a mask token for discrete states. Right depicts the conditional sampling of $x_t|Z$, where elements can split and be deleted but only, in the conditional path, according to the pre-determined branching structure of the two trees $\mathcal{T}$ in $Z$.} \label{fig:construction}
\end{figure*}

\section{Branching Flows}
In Generator Matching, conditioning on $Z$ ensures that the conditional process terminates at a sample from the data distribution, and can be used to control the properties of the conditional paths (e.g. `minibatch optimal transport' flow matching, \citet{tong2024improvinggeneralizingflowbasedgenerative}, which samples from coupled $X_0$ and $X_1$ jointly). With Branching Flows, we make \emph{extensive} use of $Z$, which we first describe in section \ref{sec:Z}, and then describe the conditional paths given $Z$ in section \ref{sec:condpaths}.

\subsection{$Z$}
\label{sec:Z}

At time \(t\) the state is an ordered list $X_t=(X_t^{(1)},\dots,X_t^{(L_t)})$  with variable length \(L_t\), where each element $X_t^{(i)}$ is a random variable on some space $\mathcal{E}$. An observation from the data distribution is $x_1=(x_1^{(1)},\dots,x_1^{(\ell_1)}) \sim q$, and a sample from the initial distribution is $x_0=(x_0^{(1)},\dots,x_0^{(\ell_0)}) \sim p$, where $\ell_0$ does not necessarily equal $\ell_1$.

We augment our sequence at $t=1$ by sampling and inserting `to be deleted' states $ø_1$, each also in $\mathcal{E}$, into $X_1$ yielding $X^{+ø}_1$, such that $X_1 = \psi_{ø}(X^{+ø}_1)$ where $\psi_{ø}$ discards all `to be deleted' states. We denote the length of $X^{+ø}_1$ as $L^{+ø}_1$, and require that $L^{+ø}_1 > L_0$ (condition 1).

To couple $X_0$ to $X^{+ø}_1$, we sample an ordered list of labeled binary rooted plane trees, $\mathcal{T} = (\mathcal{T}_1, \dots, \mathcal{T}_{L_0})$, where the root of each tree $\mathcal{T}_i$ is associated with the element $X_0^{(i)}$. Each element of $X^{+ø}_1$ is associated with a single leaf of a single tree, such that the ordering of the leaves (since the trees are planar) matches the ordering of $X^{+ø}_1$. All nodes in the tree are labeled `surviving' except leaf nodes associated with `to be deleted` states, which are labeled `deleted'. The number of descendants of node $n_i$ (including surviving and deleted) is denoted $w_i$.

To steer the conditional process along internal branches, we introduce `anchors' $\mathcal{A}$, where each node's anchor $A_i$ is a random variable on $\mathcal{E}$, requiring the anchor of the surviving leaves to equal the leaf-associated element of $X^{+ø}_1$ (condition 2). Internal anchors $\mathcal{A^I}$ have no such constraint, and can e.g. be chosen to steer the process along internal branches towards the centroid of the descendant elements.

For Branching Flows, $Z\sim p_Z$ is a tuple $(X^{+ø}_1, X_0, \mathcal{T}, \mathcal{A})$, which is obtained by first sampling $X_1 \sim q$, and then sampling from $p(x^{+ø}_1, x_0, \mathcal{T}, \mathcal{A} \mid X_1=x_1)$ according to a scheme that satisfies conditions 1 and 2 above, and that also admits a simple marginal $p(x_0)$. Note that, since $\mathcal{T}$ carries labels for which leaves are deleted, $X_1$ is recoverable from $Z$. To sample \(Z\) in practice, to be able to ensure that conditions 1 and 2 are met, we factorize $p(x^{+ø}_1, x_0, \mathcal{T}, \mathcal{A} \mid x_1)p(x_1)$ as shown in fig. \ref{fig:construction}. See App.~\ref{app:z-sampling} for concrete schemes for sampling $Z$, especially $\mathcal{T}$ and $\mathcal{A}$.

\subsection{Conditional Paths}
\label{sec:condpaths}
We construct conditional paths where elements evolve along branches of our trees, splitting and deleting, to terminate at $X_1$. We need to specify i) how the elements evolve along a branch, ii) what happens when they split, iii) the per-element split rate, and iv) the per-element deletion rate.

Between split or deletion events, each element evolves independently of the other elements along the branch with which it is associated. This is governed by a `base' generator, which can be any element-wise conditional process generator from Generator Matching, allowing drift, diffusion, and jumps, and conditioned to terminate at a specific value by $t=1$. In Branching Flows, the anchor associated with each branch steers the base generator, which is conditioned to terminate, were it to reach $t=1$, at the anchor $a_i$ of the node ahead of it.

When an element is split it is replaced by two adjacent duplicates, and each duplicate is associated with a child branch. To control the rates of split events, let \(H_\mathrm{split}\) be a `split hazard' distribution supported on \([0,1]\) with no atom at $1$, with its associated hazard rate $h_\mathrm{split}(t) = \frac{f_{H_\mathrm{split}}(t)}{1-F_{H_\mathrm{split}}(t)}$ where $f_{H_\mathrm{split}}(t)$ is the density at $t$ and $F_{H_\mathrm{split}}(t)$ is the cumulative distribution.
For an element associated with a specific branch, the tree (which is pre-sampled, within $Z$) determines the number of remaining split events $ w_i - 1$ to be realized by $t=1$.
The instantaneous per-element split event rate is $(w_i-1) \cdot h_{\operatorname{split}}(t) $, and it holds that all remaining splits are consumed by $t=1$ when $w_i >1$.

For deletions, let \(H_\mathrm{del}\) similarly define $h_\mathrm{del}(t)$, and the deletion event rate is $h_{\operatorname{del}}(t)$ on branches associated with deleted leaves, and $0$ on branches associated with internal and surviving terminal nodes.
Since the hazard rates are designed such that all splits and all deletions implied by $\mathcal{T}$ will be consumed by $t=1$, and since the anchors $a_i$ of all surviving terminal nodes are equal to the elements of $x_1$, we thus have a conditional path $x_t \sim p_{t\mid Z}(x_t\mid z)$ that begins at $x_0$ and terminates at $x_1$ with probability \(1\).

For each element to track which branch on which tree it belongs to (which is required for the conditional process to be Markov in the state), we augment the state space with a branch indicator $g_t$. With each element in $\tilde{\mathcal E}$, we have $\tilde X_t := (X_t, G_t)  \sim p_{t|Z}(\mathrm{d}x,\mathrm{d}g|z)$ that begins at $\tilde x_0:=(x_0,g_0)$ and terminates at $\tilde x_1 := (x_1,g_1)$. See Appendix \ref{app:branch-tracking} for details of our branch tracking and split and deletion operators, and see figure \ref{fig:construction} for a depiction of $Z$ and the conditional path construction.


\subsection{Sampling from the Conditional Path}

For training, we need samples at time $t$ from the conditional path $\tilde X_t \sim p_{t\mid Z}$. With Branching Flows, even though the element evolution along a branch for the conditional paths is independent of the other elements given $Z$, because of the branching structure the distribution of the element \emph{value} at any given time is not. This prevents us from adopting the usual strategy for flow matching which draws conditional samples for each element independently. However we can efficiently sample the waiting time until the next split or deletion event (see Corollary \ref{app-corr:efficientsampling} and Remark \ref{remark:deltimes}), so if we additionally have an efficient per-element sampler for $p(\tilde x_v^{(k)} \mid \tilde x_s^{(k)},a_k)$, where $s<v$ and $a_k$ is the anchor associated with the node ahead of $\tilde x_s^{(k)}$, then we can sample from the conditional path at $t$ by recursively sampling the waiting time until the next split or deletion event, and then sampling the value of the element at the next split event (or at $t$ if the next split event is after $t$). This is, like standard generator matching's full element-wise factorization, $\mathcal{O}(L^{+ø}_1)$.

For the base process in our empirical examples below, we use an Ornstein–Uhlenbeck (OU) like process with time-inhomogeneous diffusion, and with mean-reversion towards the anchors, and for discrete states we use an instance of Discrete Flow Matching (DFM) corresponding to equation 10 in \citet{Gat2024DiscreteFM}, which includes unmasking and uniform noise. See app. \ref{app:baseproc} for details.

\subsection{Parameterization of the Generator and Loss}

The Conditional Branching Flows (CBF) loss (cf. Appendix~\ref{app:branchingflowsloss}), which uses the conditional process to train a model that samples from the marginal path, is the sum of three Bregman divergences: one for the base process, one against the split intensity, and one against the deletion intensity:
\begin{align*}
    L_{\mathrm{CBF}}(\theta) = & \mathbb E_{t\sim \mathrm{\mathcal{D}[0,1]},Z\sim p_Z,(X_t,G_t)\sim p_{t|Z}(d\tilde x|z)}\\
& \Big[ D^{\mathrm{split}}_{t, X_t}(R_{t}^{Z,G_t}(X_t), R_{t}^{\theta}( X_t))\\
& +D_{t, X_t}^{\mathrm{del}}(\rho_{t}^{Z,G_t}(X_t),\rho_{t}^{\theta}( X_t))\\
& +  D_{t, X_t}^{\mathrm{base}}\big(F_{t}^{Z,G_t,\mathrm{base}}( X_t),F_{t}^{\theta,\mathrm{base}}( X_t)\big)
\Big].
\end{align*}

Concretely, we use an elementwise loss where we parameterize the split and deletion rates in terms of the number of splits by $t=1$ and the probability of deletion by $t=1$ (i.e. ``$X_1$-prediction''), denoting $R_{t,i}^{Z,G_t}( X_t)$ for the number of split increments ahead of the $i$'th component, and $\rho_{t,i}^{Z,G_t}( X_t) \in \{0,1\}$ for the $i$'th component of $\rho_t^{Z,G_t}( X_t)$. These are parametrized by $R_{t,i}^\theta( X_t) \in (0,\infty)$ and $\rho_{t,i}^\theta( X_t) \in (0,1)$ respectively. We write $F_{t,i}^{Z,G_t,\mathrm{base}}(X_t)$ as the `base' generator of a per-element conditional process along a single branch, steered to terminate (by $t=1$) at the anchor $a(i,Z,G_t)$. This is the anchor at the end of the branch along which the $i$-th component of $X_t$ is evolving (tracked by the branch indicator of the $i$-th component of the augmented state $\tilde X_t$). Our loss is then:\\
\begin{equation}\label{eqn:concreteusedloss}
    \begin{aligned}
        &  L_{\mathrm{CBF}}(\theta) = \mathbb E_{
t\sim \mathrm{\mathcal{D}[0,1]},
Z\sim p_Z,
 (X_t,G_t)\sim p_{X_t,G_t|Z}(\mathrm{d}x,dg|z)
}\\
&\qquad\Big[\sum_{i=1}^{L_t}\Big(
{R_{t,i}^\theta( X_t)-R_{t,i}^{Z,G_t}(X_t)\log R_{t,i}^\theta(X_t)}\Big)\\
&\qquad+\sum_{i=1}^{L_t} \Big(- [\rho^{Z,G_t}_{t,i}(X_t)\log \rho_{t,i}^\theta(X_t) + (1-\rho^{Z,G_t}_{t,i}(X_t))\log(1-\rho_{t,i}^\theta(X_t))]\Big) \\
&\qquad+ \sum_{i=1}^{L_t} \Big(D_{t,X_t,i}^{\mathrm{base}}(F_{t,i}^{Z,G_t,\mathrm{base}}(X_t),F_{t,i}^{\theta,\mathrm{base}}( X_t))
\Big)\Big].
\end{aligned}
\end{equation}

Note that only $X_t$ (and not $G_t$) is input to the model, as $G_t$, like $Z$, is marginalized out in the training process per app. \ref{app:auggm}.

\subsection{Training and Sampling}
Training proceeds exactly as is standard in flow matching, by stochastic gradient descent. We draw $t \sim \mathcal{D}$, $X_1 \sim q$, $Z \sim p_{Z|X_1}$, and then draw $\tilde X_t = (X_t,G_t) \sim p_{t\mid Z}(d\tilde x_t|z)$. From $Z, X_t$, and $G_t$, we recover $R^{Z,G_t}_{t,i}( X_t)$, $\rho^{Z,G_t}_{t,i}( X_t)$, and $F^{Z,G_t,\mathrm{base}}_{t,i}( X_t)$, and take gradient steps on the loss in equation \eqref{eqn:concreteusedloss}.

We sample from the marginal paths by taking small steps in time. A trained model outputs element-wise predictions $R^\theta_{t,i}( X_t)$, $\rho^\theta_{t,i}( X_t)$, and $F^{\theta,\mathrm{base}}_{t,i}( X_t)$. For a step size $\Delta t$, we first take a step on the base process according to the model's learned expected generator, independently per element, just as in standard Generator Matching. For the branching and deletion part of each step, we calculate per-element split intensities $h_{\operatorname{split}}(t) \cdot R^\theta_{t,i}( X_t)$ and deletion intensities $h_{\operatorname{del}}(t) \cdot \rho^\theta_{t,i}( X_t)$, and we sample which elements split or delete within the step. Elements with deletion events are removed, and elements with split events are duplicated in-place. The trees and branch indicators have been marginalized out and are not considered when sampling.

\begin{figure}[t!]
\centering{\includegraphics[width=0.65\columnwidth]{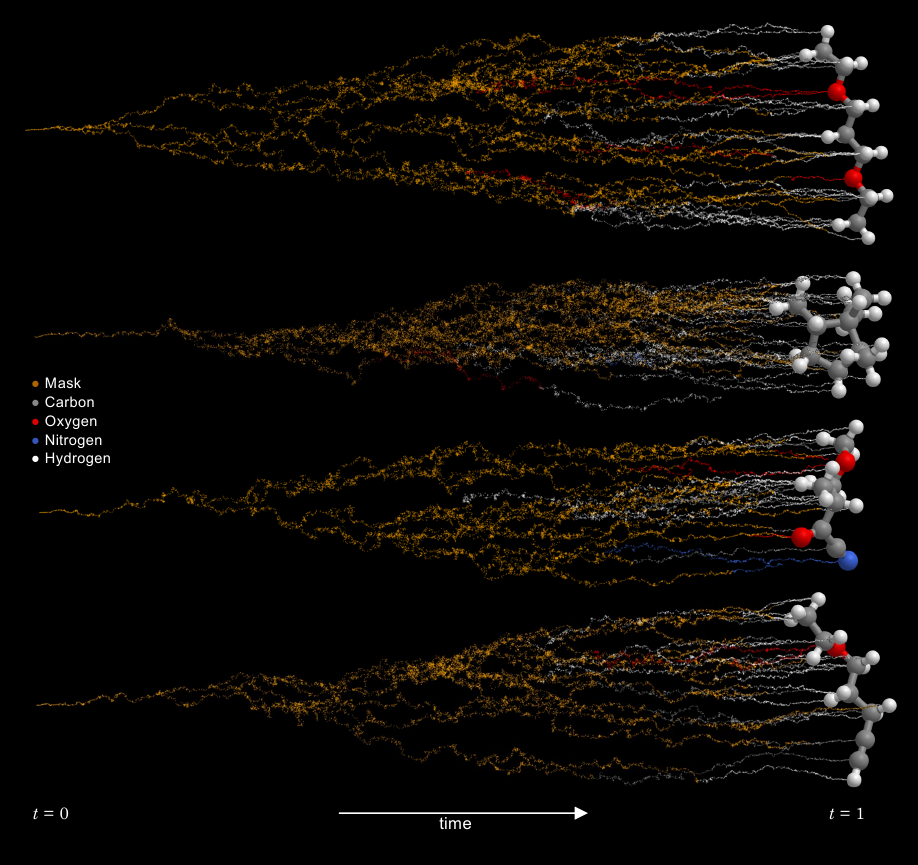}}
    \caption{\textbf{QM9 Branching Flows sampling trajectories.} A visualization of inference trajectories from a QM9-trained Branching Flows model. The final sampled molecule, when $t=1$, is depicted on the right. At every inference step from $t=0$ to $t=1$, we draw the current $x_t$ state as colored points, with a small rightward displacement, which shows the branching and deletion history as $x_t$ is transported from a single element when $t=0$ to the final sampled molecule. The color of the trails shows the atom type which begins as `Mask' (in orange) and switches to concrete atoms as $t \rightarrow 1$.}
    \label{fig:QM9path}
\end{figure}

\section{Results}
In order to evaluate the performance of Branching Flows on situations where length variation is crucial we consider the following four tasks: i) Atom position and type generation for small molecules (continuous and discrete), ii) Antibody heavy chain generation (discrete sequence only), iii) Protein backbones, comprising amino acid backbone `frames' \citep{jumper2021highly} with Euclidean positions, \(SO(3)\) orientations and discrete amino acid labels (continuous, manifold and discrete) iv) Partially-latent all-atom  protein structure and sequence co-design (continuous).

\subsection{QM9}

\begin{figure*}[b!]
\centering{\includegraphics[width=0.99\textwidth]{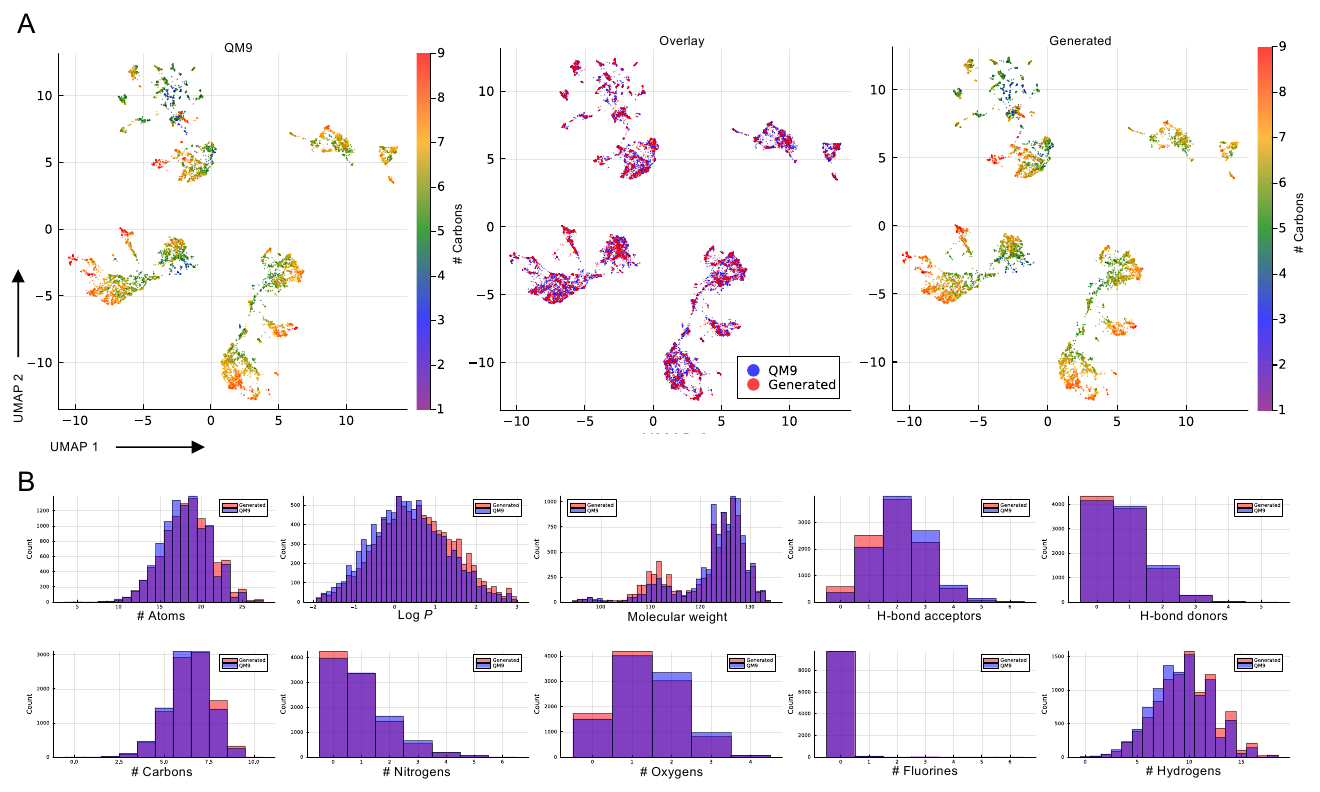}}
    \caption{\textbf{QM9 data vs Branching Flows samples.} Panel A computes molecular fingerprints (using RDKit) and jointly embeds QM9 data and generated samples using UMAP. Left shows QM9-only embeddings, colored by the number of carbon atoms in each molecule; right shows generated samples, similarly colored, and center shows the overlap of QM9 (blue) vs Branching Flows generated samples (red), showing similar distributions. Panel B depicts property distributions from QM9 and Branching Flows generated samples, including atom counts, as well as a number of standard molecular descriptors computed with RDKit.}
    \label{fig:QM9stats}
\end{figure*}

\begin{figure*}[h]
\centering{\includegraphics[width=0.99\textwidth]{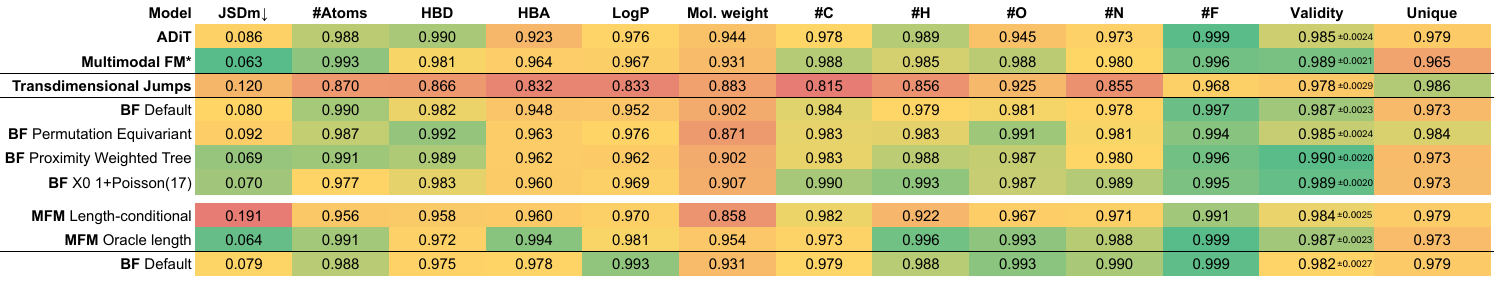}}
    \caption{\textbf{QM9: model behavior.} Here we compare samples generated by Branching Flows (including a `Default' variant, and a number of variants described in App \ref{SI:variations}). The top 7 rows show unconditional molecule generation. We compare to fixed-length baselines ADiT \citep{joshi2025allatomdiffusiontransformersunified} and an architecture, process, and training matched fixed-length Multimodal Flow Matching (MFM) variant, which are both provided with the true length, and then with a variable length Transdimensional Jump Diffusion \citep{Campbell2023TransDimensionalGM}. 
    The bottom three rows show results from a `molecule completion' task where part of the molecule is fixed. Here we compare to a matched Oracle Length MFM model (note: this has length information that would be unavailable for `completion' tasks) and an MFM model that attempts to draw the completion length so that the marginal length distribution is matched.
    Distribution-matching comparisons include a binned multivariate Jensen-Shannon divergence (JSDm), and 1-KS statistics (used for columns 2 to 10).}
    \label{fig:QM9maintable}
\end{figure*}

\paragraph{QM9.} QM9 \citep{ramakrishnan2014quantum} contains small organic molecules selected from the GDB (Generated Database) chemical space enumeration. We train Branching Flows to jointly generate atom coordinates and discrete labels; full data preprocessing, Branching Flows specification, and model/training details are in Appendix~\ref{dat:qm9}. Figure \ref{fig:QM9stats} compares generated and data distributions (10k samples each) for molecule fingerprints, which encode molecule properties into a fixed-length bit vector, embedded into two-dimensions by UMAP \citep{mcinnes2020umapuniformmanifoldapproximation}, showing no obvious differences. Also shown are histograms of atom counts and molecule properties, with some subtle departures, most visible in $\operatorname{LogP}$ and molecular weight. Using a Kolmogorov-Smirnov based statistic ($1-KS_D$; see App. \ref{dat:qm9}) to quantify property agreement between samples and the data distribution (Figure \ref{fig:QM9maintable}), we also compare 10k samples from a fixed length oracle (matched in architecture, process, and training) which is provided with the correct sequence length at training time, and a sample from the data length at generation time;  a fixed-length baseline ADiT \cite{joshi2025allatomdiffusiontransformersunified}; and from a QM9-trained `Transdimensional Jump' diffusion model \citep{Campbell2023TransDimensionalGM}.

Further, we use QM9 to explore the behavior of Branching Flows across the design space, modifying the way the trees and anchors are sampled, modifying the $X_0$ size distribution, ablating splits, ablating (and oversampling) deletions, modifying the atom ordering, and considering fully permutation invariant model variants. As shown for a subset of models in Figure \ref{fig:QM9stats} and a larger set in Appendix Fig. \ref{fig:variations}, performance is robust across this large theoretically-valid design space, except where choices are made to be intentionally challenging for the model (e.g. randomized atom ordering). Further, the large design space allows domain-respecting choices that provide minor performance advantages.

We also consider a `molecule completion' task, where some atoms are fixed (see Appendix section \ref{qm9-app:qm9modelandtraining}). In this instance, the Oracle length model has information that simply isn't available at sampling time, because, conditioned on a subset of the molecule, the number of remaining atoms for each designable portion of the molecule isn't known. Here, Branching Flows again performs close to this oracle, and outperforms (especially in distribution matching) a fixed-length model that attempts to use a simple heuristic length sampling strategy.

\subsection{Antibody Sequences}

\begin{figure}[t!]
    \centering{\includegraphics[width=0.99\textwidth]{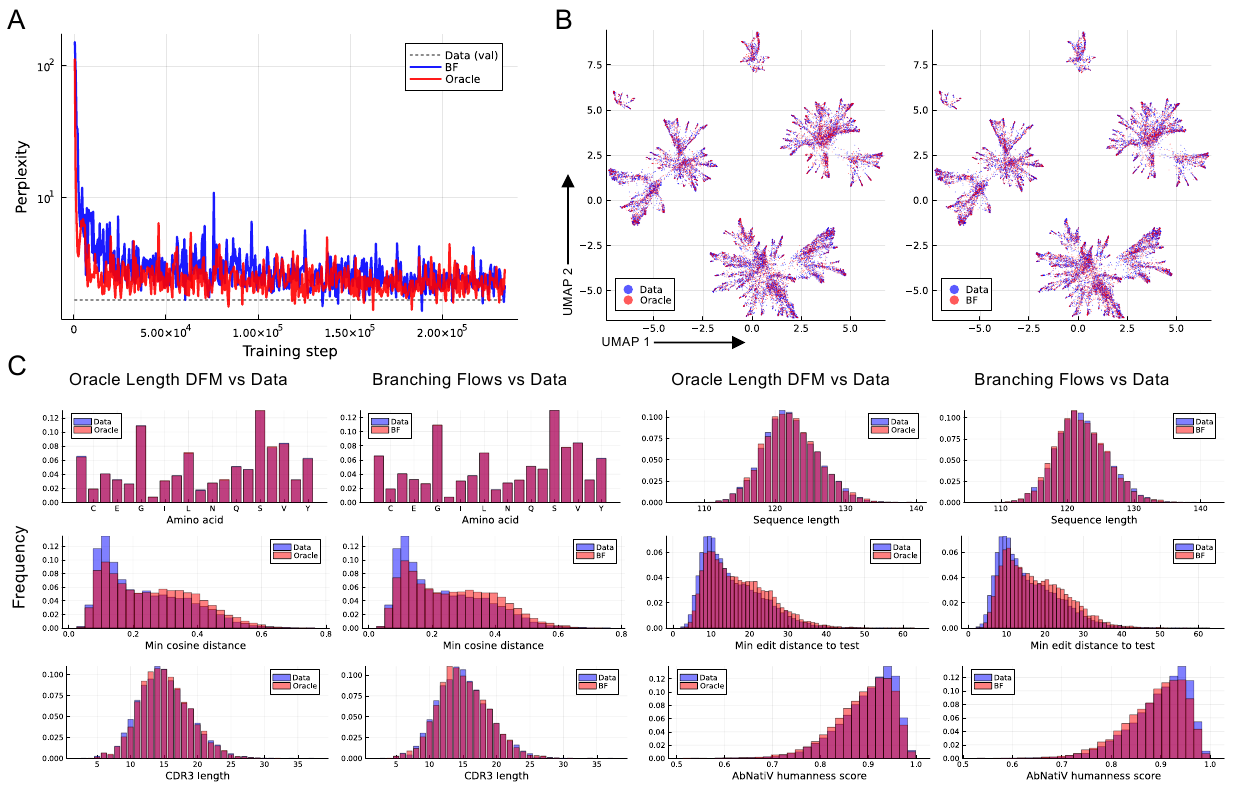}}
    \caption{\textbf{Antibody generation distribution matching.} 
    \textbf{A} The perplexity, evaluated under an autoregressive LLM, of samples generated by Branching Flows vs an oracle-length discrete flow matching model (where the length for each sample is taken from a random real sequence). This is shown over training iterations. \textbf{B} two seqUMAP \citep{hanke2022multivariate} plots comparing the clustering of real data with the oracle-length samples and the Branching Flows samples, respectively. \textbf{C} Comparison of several distributions of the generated sequences with sequences from the validation data set.
    }
    \label{fig:abstats}
\end{figure}

We investigate a discrete sequence learning task: antibody amino acid sequence generation. We compare a Branching Flows model to an `oracle-length' discrete flow matching model, which is trained knowing the true length, and sampled using a `true' length from the data distribution. We also compare to a re-implementation of Edit Flows \citep{Havasi2025EditFF}. Dataset details, Branching Flows specification, and model/training details are in Appendix~\ref{dat:abs}. Figure \ref{fig:bf_vs_ef} (Appendix) shows the distributions of statistics (App. \ref{dat:abs}) from sequences generated by both Branching Flows and Edit Flows models, perplexity-over-time training dynamics, and seqUMAP \citep{hanke2022multivariate} representations of the $10\,000$ generated sequences. Overall, Branching Flows is close to the oracle length model, can learn sequence-position-specific amino acid frequencies (Figure \ref{fig:abpos}, Appendix), and outperforms Edit Flows on property matching (Table \ref{table:abcomptable} and Figure \ref{fig:bf_vs_ef}, Appendix).

Similar to QM9, we also consider a conditional `completion' task for antibody sequences, where the model is required to fill in missing segments when their length is unknown. Here we similarly outperform Edit Flows, with performance closer to the Oracle Length model (which, again, uses information not usually available at inference time).

\begin{table}[t]
\caption{Distribution agreement ($1-\mathrm{KS}_D$, higher is better) between properties of samples from each model vs.\ validation sequences. From left to right: Sequence Length, Diversity (within-samples min Cosine distance), Novelty (minimum edit distance to true sequence), CDR3 length, and AbNatiV `humanness' score. Histograms of these properties are shown in Figure \ref{fig:abstats}. Shown are results for an unconditional task where the entire sequence is generated, and a conditional task that generates multiple missing segments of the sequence. For the conditional task the `Oracle length' model knows the length of each missing segment.}\label{table:abcomptable}
\centering
\small
\begin{tabular}{llccccc}
\toprule
& \textbf{Method} & \textbf{Length} & \textbf{Diversity} & \textbf{Novelty} & \textbf{CDR3 length} & \textbf{AbNatiV} \\
\midrule
\multirow{3}{*}{\small{\textit{Unconditional}}}
& Oracle length    & 0.995 & 0.899 & 0.903 & 0.984 & 0.943 \\
& Branching Flows  & 0.986 & 0.894 & 0.888 & 0.989 & 0.931 \\
& Edit Flows       & 0.989 & 0.713 & 0.770 & 0.983 & 0.769 \\
\midrule
\multirow{3}{*}{\small{\textit{Conditional}}}
& Oracle length    & 0.986 & 0.972 & 0.956 & 0.987 & 0.988 \\
& Branching Flows  & 0.976 & 0.942 & 0.917 & 0.978 & 0.981 \\
& Edit Flows       & 0.931 & 0.885 & 0.869 & 0.968 & 0.886 \\
\bottomrule
\end{tabular}
\end{table}

\subsection{Protein Structure and Sequence}\label{protsec}

\subsubsection{SE(3) equivariant backbone plus discrete sequence}

\begin{figure*}[b!]
    \centering{\includegraphics[width=0.99\textwidth]{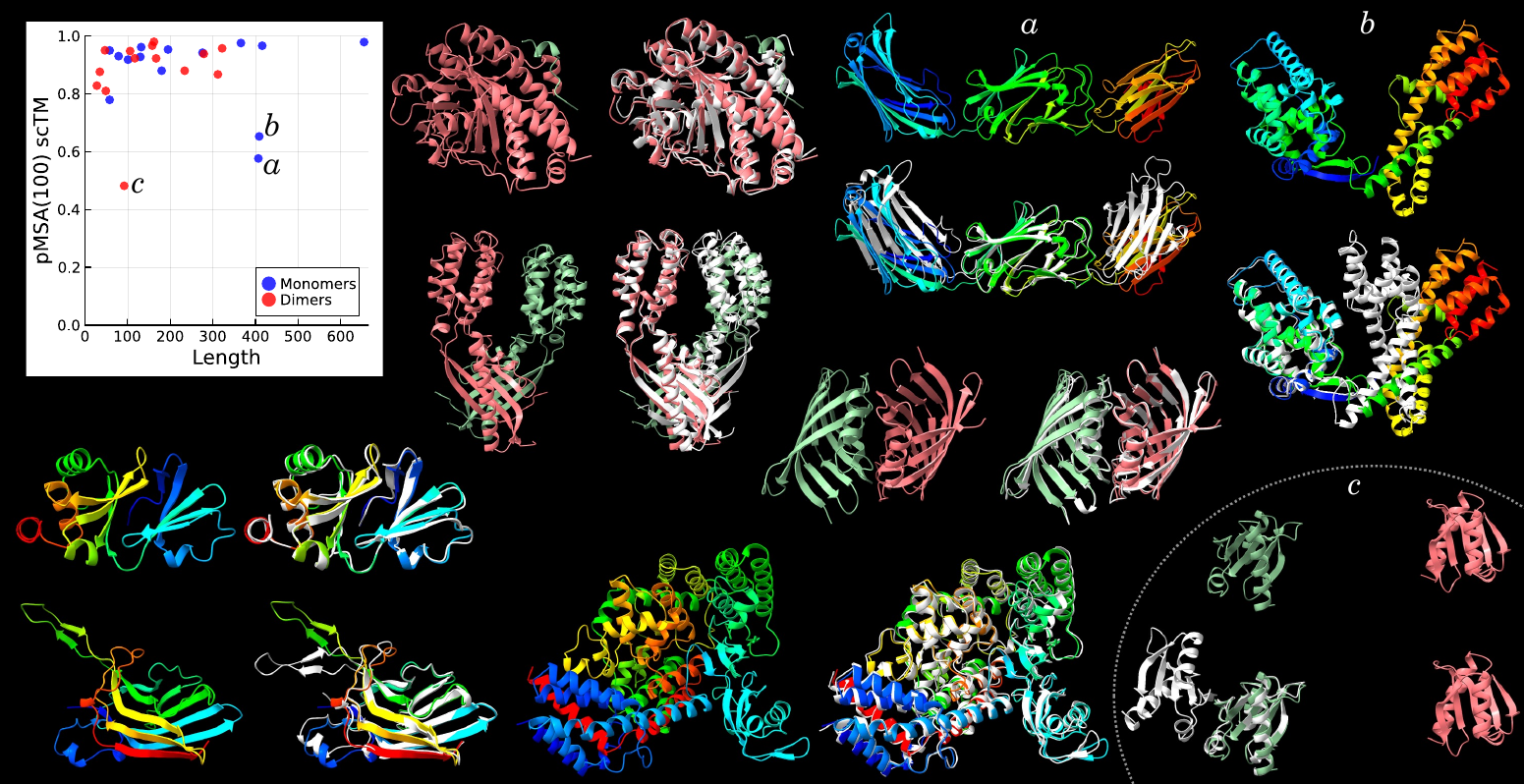}}
    \caption{\textbf{Branching Flows protein samples.} Top left inset shows pseudoMSA scTM refolding scores, and selected samples are shown (rainbow colored for single chains, and per-chain for dimers). To the right or below each is shown an overlay of the sample and the refolded structure (grey). Specifically, the three worst scTMs are depicted (labelled a, b, and c on the inset and the main panel) showing that the low scTMs are driven by unpredictability due to flexible linkers, or incorrect dimerization prediction.}
    \label{fig:protsamps}
\end{figure*}

\begin{figure*}[t]
    \centering{\includegraphics[width=1.0\textwidth]{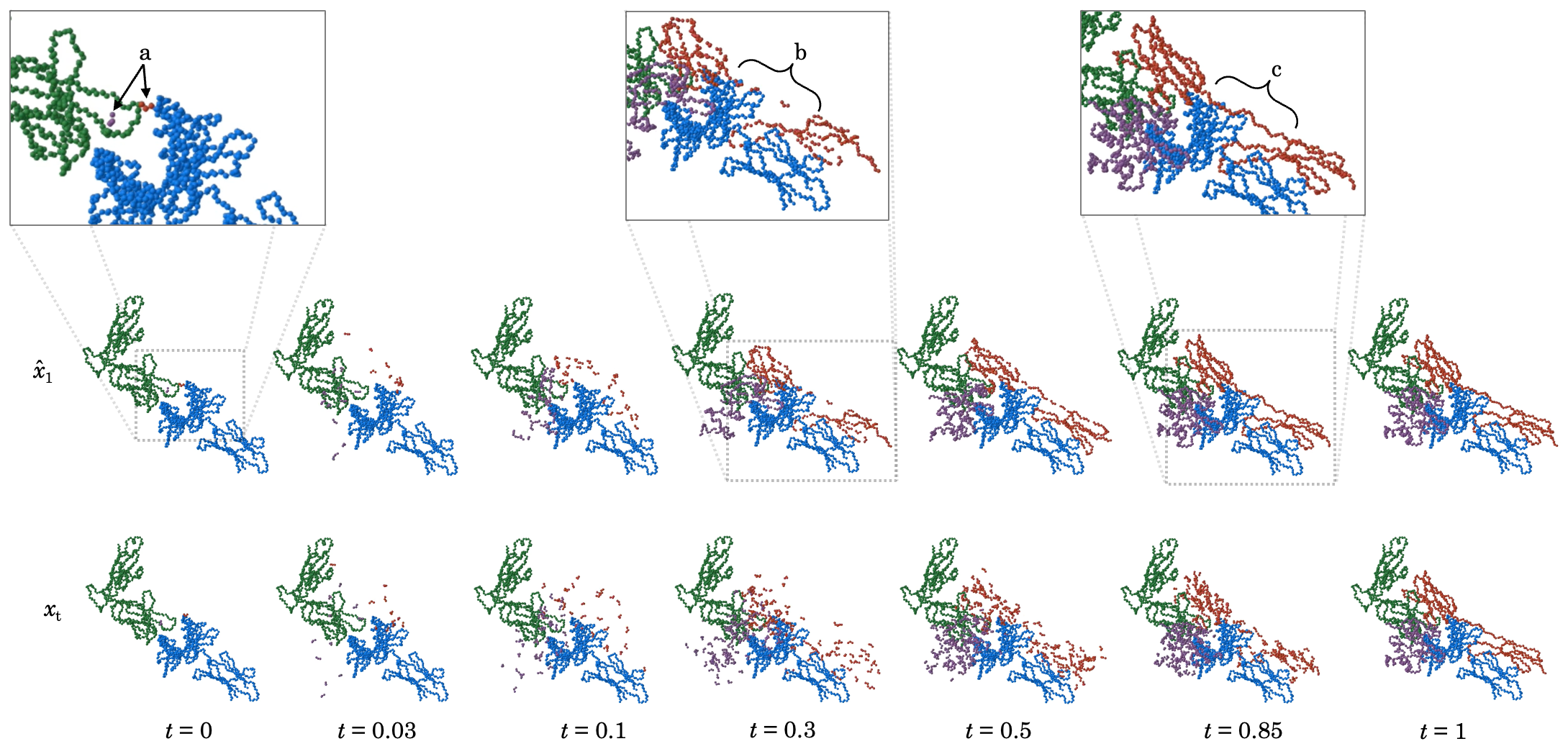}}
    \caption{\textbf{Branching Flows protein sampling trajectories.} Shown are snapshots from a sampling trajectory from BF-ChainStorm, with both the current state $x_t$ and the model's prediction of the end state, $\hat{x_1}$. The trajectories are colored by chain, and each backbone residue is shown as three spheres $(N,C\alpha, C)$. Here, two chains were fixed, and two chains were designable, with each designable chain starting from a single residue (arrow a in the first inset). The chains grow over time, and in this example the orange chain converges on two separate domains. Interestingly, during the intermediate flow state ($t=0.3$, bracket b in the second inset), as the two domains resolve in $\hat{x_1}$, they are not connected to each other, and the model uses split events to build a complete linker between them by $t=1$.}
    \label{fig:prottraj}
\end{figure*}

\begin{figure*}[t]
    \centering{\includegraphics[width=1.0\textwidth]{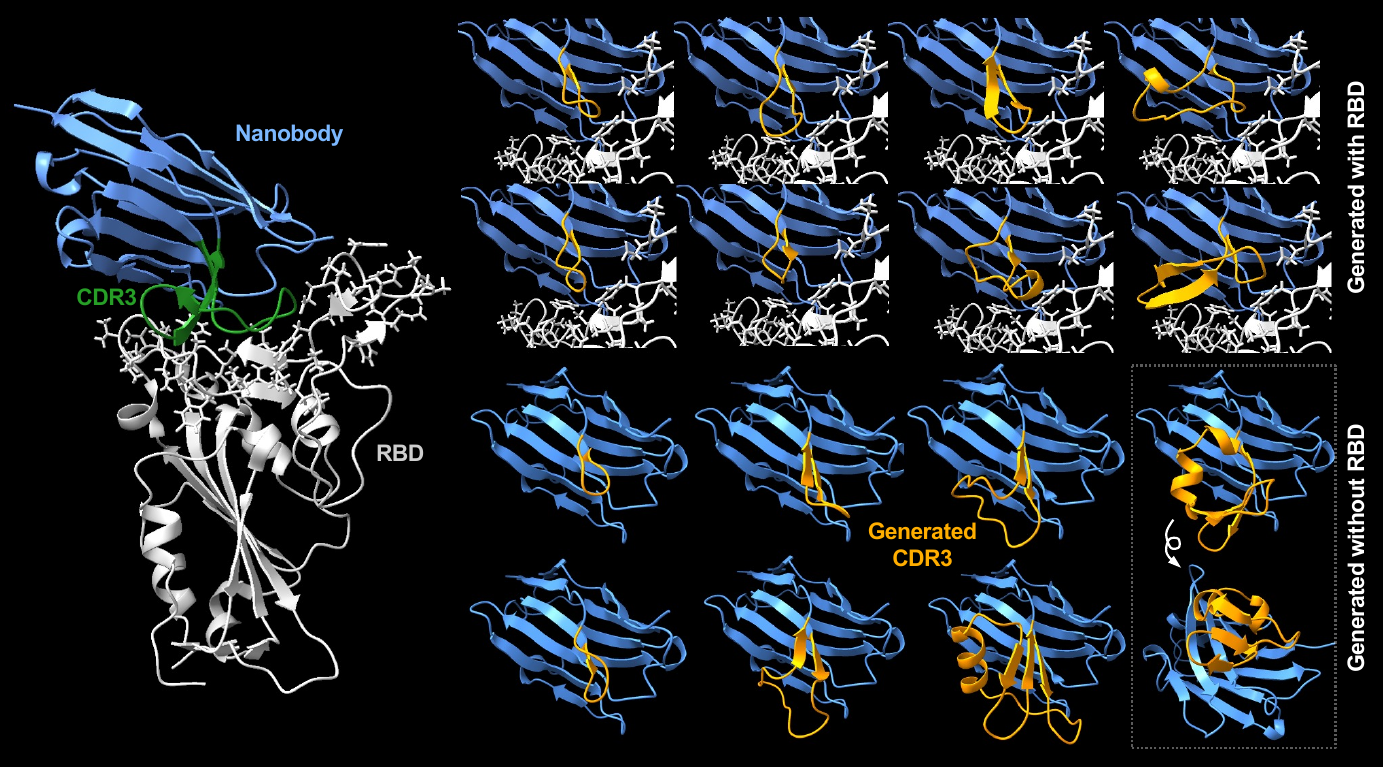}}
    \caption{\textbf{Branching Flows protein `infix' samples.} Depicted is the template PDB (9IQP, \cite{shcheblyakov2025ultra}) where only the CDR3 (green in the template) was designable, and the rest of the structure was fixed and conditioned upon. The top two rows of samples show generated CDR3s when they are generated in the context of the binding partner, and the bottom two rows show generations in the absence of the binding partner.}
    \label{fig:protinfix}
\end{figure*}

Starting with ChainStorm \citep{OrestenChainStorm}, a multimodal flow matching model that jointly generates protein backbones and amino acid sequence labels, as a base model, we finetuned two variants: For `BranchChain', the designable residues are complete chains, randomly selected for masking during training, so it can generate chains unconditionally or it can generate chains conditioned on other fixed chains. BranchSegment's conditioning mask during training is a random number of random-length contiguous segments, so multiple segments can be simultaneously designed during inference. Full setup and evaluation details are in Appendix~\ref{dat:prot}.

To investigate whether a Branching Flows protein model can generate samples with reasonable protein-like geometry, we sampled, without conditioning, 20 monomers and 20 dimers, discarding samples with chain breaks (one monomer and five dimers) and samples too short for meaningful evaluation ($<25$ residues in the longest chain: five monomers, and 1 dimer). Using the `pseudoMSA refolding' strategy  \citep{OrestenChainStorm}, we generated a pseudoMSA of 100 sequences for each design with ProteinMPNN \citep{dauparas2022robust}, and evaluate the `self-consistency TM scores' of a single refold, via Boltz2 \citep{passaro2025boltz} from this pseudoMSA.

Figure \ref{fig:protsamps} shows the scTM scores, most very close to 1, showing that our designs have geometry consistent with Boltz2. We also depict selected examples, for monomers and dimers, of BranchChain generated proteins and their refolds, specifically showing the three cases with low scTMs. For the two poorly refolded monomers, this is because the generated structure had one or more flexible linkers, and the individual domains were well refolded. For the poorly refolded dimer, this was because BranchChain generated two disconnected chains, and Boltz2 placed these adjacent. Overall, BranchChain generates samples with a variety of protein chain lengths, with plausible backbone geometry. Further, as can be seen in many of the samples in figure \ref{fig:protsamps}, the propensity of the ChainStorm base model to sample homodimers is preserved. This is interesting given that ChainStorm's single attention head that governs symmetric oligomer sampling is primarily driven by chain lengths which are fixed in ChainStorm, but now stochastically generated in BranchChain.

For illustration purposes, we also finetuned a BranchChain variant with only a single $x_0$ element per chain. Figure \ref{fig:prottraj} shows steps along the sampling trajectory for a sample where two chains (blue and green) were fixed, and two chains were sampled. The orange chain, best seen through the lens of the model-predicted end state $\hat{x_1}$, highlights a key Branching Flows feature: at first, the designed chain converges upon two separate domains, with insufficient residues in between them for them to be linked. This is then progressively filled in by insertions, until a complete linker connects them as a single chain.

For BranchSegment we do not attempt quantitative evaluations, but we demonstrate that it is a promising solution to the `unknown length infix sampling' problem. Taking a protein complex (PDB: 9IQP, \citet{shcheblyakov2025ultra}) of a nanobody and the SARS-CoV-2 receptor binding domain (RBD), we mask only the `Complementarity Determining Region 3' (CDR3) as designable, and we generate 20 samples from the RBD-bound complex, and 20 samples from the unbound nanobody alone. The sampled CDR3s had a broad length distribution (7 to 29 residues for bound, 7 to 40 unbound). For the bound samples, even though the side chain atom positions from the RBD were not provided to BranchSegment (it only sees backbone frames and amino acid labels) the designed CDR3s had no atomic clashes with the RBD side chain rotamers for 15 out of 20 of the generated CDR3s. Some of the unbound CDR3s appear structurally plausible, but are extremely unlikely to occur naturally, which is not surprising since neither the base model nor BranchSegment have any antibody or nanobody-specific training (see fig \ref{fig:protinfix}).

\begin{table}[b]
\caption{\textbf{Branching La Proteina metrics.} Designability, co-designability, and diversity (structure only, and structure \& sequence) for $N=500$ samples. Intervals are symmetric two-sigma Wald intervals for binomial proportions.}
  \centering
  \begin{tabular}{lcccc}
  \hline
  Sample set & MPNN1 designability & Co-designability & Div. str & Div. str+seq \\
  \hline
  Branching Flows LD3
  & $0.990 \pm 0.009$
  & $0.966 \pm 0.016$
  & 128 & 215 \\
  La Proteina LD3
  & $0.972 \pm 0.015$
  & $0.952 \pm 0.019$
  & 110 & 173 \\
  \hline
  \end{tabular}
  \label{laprotmetrics}
\end{table}

\subsubsection{Partially-latent Protein Structure}
La Proteina \citep{geffner2025laproteinaatomisticproteingeneration} uses "partially-latent" Flow Matching where each residue in a protein chain is represented by its C-alpha coordinates, and an 8-dimensional latent encoding the residue label and the placement of other atoms, obtained via a Variational Autoencoder. We finetune Branching Flows into this by introducing split and deletion heads into the LD3 La Proteina model. Table \ref{laprotmetrics} shows that the Branching Flows finetune achieves excellent designability (via an MPNN-generated sequence), co-designability (sequence via the sampled latents), and diversity metrics.

\section{Related Work}

\subsection{Edit Flows} 
\citet{Havasi2025EditFF} describe Edit Flows, a method for varying-dimensionality generation of discrete sequences. Edit Flows works by parameterizing the rates of \textit{edit operations} (insert, delete and substitute). During training, an \textit{auxiliary alignment process} is introduced. This entails adjoining to the alphabet a null token \(\varepsilon\) such that the sequence length of \(X_0\) is the same as the sequence length of \(X_1\). One denotes this new space by \(\X \times \Zaux\). Then, transitions in this augmented space are interpreted as either insertions (swapping \(\varepsilon\) to a token in the alphabet), deletions (swapping a token in the alphabet for \(\varepsilon\)) or a substitutions (swapping two tokens in the alphabet). The model only sees $t$ and \(\X\), and the marginal rates of a CTMC on \(\X\) are learnt by training against a CTMC on \(\X \times \Zaux\). Edit Flow's main contribution is proving, for the discrete case, that you can marginalize over this time-dependent latent process. In Branching Flows we sample a non-time-dependent latent that fully specifies how elements in \(X_0\) will map, via splits or deletions, to elements in \(X_1\), allowing us to extend beyond discrete states, and affording flexibility in how the sample paths are constructed. Furthermore, to extend Edit Flows to a continuous setting one would have to parametrize and learn a flexible distribution over insertions into a continuous space, which is often hard in practice. Later, One Flow \citep{OneFlowNguyen} uses the approach of Edit Flows to allow the interleaving of images and text by inserting `image tokens'.

\subsection{Transdimensional jumps}
Transdimensional jump diffusion \citep{Campbell2023TransDimensionalGM} models continuous-valued variable-cardinality states. Between jumps, a base SDE evolves the continuous coordinates, while at jump times, the dimension changes by one. In the forward direction (data noising --- diffusion models adopt the opposite convention to flow matching), one component is deleted at a specified rate, chosen such that the forward process will terminate at a single surviving component.

The time reversal of the forward process is again a jump diffusion process, but where elements are inserted into the state, and a neural network parametrizes the insertion state and post-insertion distribution of the inserted element. One challenge is that, unlike Branching Flows's `split' mechanism, the insertion distribution does not have a simple form, and any tractable approximation used in practice (e.g. a Gaussian) can cause a mismatch between the conditional and marginal process.

\subsection{DrugFlow}  
In \citet{Schneuing2025MultidomainDL} a trans-dimensional model for small molecule generation is presented. It works by introducing a virtual atom type whose coordinates are set to the center of mass of the ligand and which have no bonds. During training the training data is augmented by adding \(n_{\text{virt}} \sim U(\{0,1,\dots,N_{\text{max}}\})\) virtual nodes, to allow the model to learn to set some atom categories to the virtual category. The model is thus deletion-only and requires specifying a priori \(N_{max}\) which leads to a fixed upper bound on generated molecule size. This contrasts to Branching Flows, which can generate arbitrary-length objects and allows both insertions and deletions. Furthermore, elements in the virtual category are retained in the flow and evolve under it (although their bond types are set to `None'), which forces model computation to always be done on the maximum number of elements for the specific molecule being generated.

\section{The Design Space of Branching Flows}

The `design space' of fixed-length element-wise Generator Matching approaches is already large (encompassing almost all diffusion and flow matching methods), and Branching Flows layers element splitting and deletion atop this, necessarily expanding the design space further.

Some choices are structurally dictated by the task. For example, if the samples are ordered sequences, then the sampled trees must be planar (ie. coalescing only adjacent elements), but if the samples are sets of elements being modelled with a permutation-equivariant model, then no such constraint is required. Further, the trees can enforce useful structural constraints: for example, if you are modeling protein sequences with multiple chains and you wish to directly control the number of chains, then during training you prevent the coalescence of any elements from different chains.

Then there are choices that are not fully dictated by the task, but where there is often a predictable inductive benefit, such as averaging for anchor choices in most continuous tasks, or preferring to merge elements that are spatially close during tree construction, both of which reduce average drift.

Just as in standard diffusion and flow matching, however, there are also choices that are not dictated by the problem structure at all, such as the hazard distributions (where mostly splits very close to $t=1$ should be avoided), and choices that involve computational tradeoffs, such as increasing the $X_0$ length, increasing the volume of deletions, or using a pure-deletion model (which all increase the $X_t$ length and thus, quadratically, the training compute).

\section{Discussion}
We have shown that Branching Flows is a capable `distribution-learner' of multimodal sequences that vary in length, addressing a previously-unsolved fundamental problem in generative modeling. For example it is, to our knowledge, the first method that can sample an infix of unknown length from a multimodal sequence. For now, we restrict ourselves to demonstrating that Branching Flows works in a variety of domains, rather than extensive comparisons and benchmarks on downstream tasks, as performance on benchmarks is often more sensitive to practically useful but theoretically uninteresting domain-specific adjustments to the model and training details, which is orthogonal to the purpose of this manuscript.

For the QM9 small-molecule dataset, we compare samples from our Branching Flows-trained model with those from the Transdimensional Jump Diffusion model of \cite{Campbell2023TransDimensionalGM}, showing favorable distribution matching for Branching Flows. We suspect this difference is attributable to two things. Firstly, transdimensional jump diffusion spawns elements at a position that is, in its conditional process, coupled to an element in the data distribution. While the conditional distribution of the spawn point is Gaussian conditioned on the corresponding $X_1$ element, the marginal distribution during inference is a (possibly infinite) mixture of Gaussians, which is challenging to approximate. Branching Flows avoids this by spawning new elements via a splitting mechanism that, conditioned on $X_t$ is independent of $X_1$. Secondly, transdimensional jump diffusion handles discrete atom types by converting them to continuous vectors. In other contexts, this does not compare favorably to discrete character diffusion models \citep{sahoo2024simpleeffectivemaskeddiffusion}.

For the antibody sequence modeling problem, we compare Branching Flows to an oracle-length model. During inference, the oracle length is sampled from the data distribution. Both the training dynamics of Branching Flows, and the ability to match the data distribution, are similar to a model that has been provided the true length. Edit Flows \citep{Havasi2025EditFF}, modeling variable length discrete sequences, is a natural comparator for the antibody sequence modeling problem. Here the model architectures are matched between Branching Flows and Edit Flows, where the only difference is that the Edit Flows insertion head is as large as the alphabet size, whereas Branching Flows only needs one splitting rate per element. While we see favorable performance of Branching Flows, one caveat is that this is our own Edit Flows implementation since, at the time of writing, there was no reference implementation provided by the authors.

As far as we are aware `BranchSegment' is the first generative model that can perform `unknown-length infix sampling' for proteins, where the backbone is constrained either side of the designable region, and the length of the designed segment is determined by the model. In practice, current generative models must use a range of pre-sampled lengths, with post-sampling filters to select from these, and non-generative models could perform sequence search on a surrogate objective (e.g. via a folding model). Neither will provide distribution-matching, and both are computationally expensive. Further, the `pre-sampled length' strategy required by fixed-length flow and diffusion models can combinatorially explode if there are multiple sequentially non-contiguous but spatially-interacting segments that need to be designed, which would be the case during protein-protein interface design or re-design.

One practically useful feature, demonstrated with `BranchChain' and `BranchSegment' is that Branching Flows can be finetuned into an existing model. Three days on a single GPU (NVIDIA RTX 6000 Ada) was sufficient. This suggests that, even though the base model was trained via fixed-length flow matching, the internal representations required to learn the branching and deletion process may already be present. This would be extremely useful if it extends to other models, allowing large models with high pretraining cost to be repurposed with Branching Flows.

If the elements are unordered, then Branching Flows can still be used, with adjustments, for example with the model being equivariant to permutations of the elements. Initial experimentation in this setting suggests that such models exhibit slower initial training, but their eventual performance is similar to those where an ordering is imposed.

Branching Flows is, to our knowledge, the first `variable length Generator Matching' method, and we anticipate that there is a large design space of alternatives, including methods that avoid the `splitting' mechanism and instead spawn elements at points in space controlled by the model, or from some fixed base distribution.  Using the QM9 dataset, we have begun a systematic exploration of model behavior across the Branching Flows design space, but this should be extended to other domains, and compared to other potential variable length Generator Matching methods.

\section{Acknowledgments}
This project received support from the Swedish Research Council (2024-00390 and 2023-02516) and the Knut and Alice Wallenberg Foundation (2024.0039) to B.M. Development of key dependencies (e.g. \url{https://github.com/MurrellGroup/Onion.jl}) used in this work was enabled by the Berzelius resource provided by the Knut and Alice Wallenberg Foundation at the National Supercomputer Centre.
AI disclosure: GPT5 was used for initial drafting of some sections, which were rewritten and checked by the authors.


\bibliography{library}
\bibliographystyle{apalike}

\appendix

\input{refactored_appendices/A_concrete_z_schemes}

\input{refactored_appendices/B_splits_deletions}
\input{refactored_appendices/C_branch_tracking}

\input{refactored_appendices/D_base_processes}

\input{refactored_appendices/E_generators}
\input{refactored_appendices/F2_analyses}

\end{document}

%% file: refactored_appendices/A_concrete_z_schemes.tex
\section{$Z$ sampling schemes}
\label{app:z-sampling}

Here we outline concrete schemes for sampling $Z := (X_1^{+ø}, X_0, \mathcal{T}, \mathcal{A})$, starting from $X_1$, that we use in our empirical examples below. Some datasets have elements that are naturally grouped (e.g. proteins, grouped by chain), and one can construct $Z$ to respect these groupings, where it makes sense to do so (see section \ref{protsec} for an example). In the case where there is only one group, all elements have the same group index. Further, if any elements are `fixed' (e.g. in the protein example below, by a conditioning mask), then any elements with a fixed element between them in the sequence must belong to a different group.

\subsection{Initial Distribution: $X_0 \sim \mathcal{D}_0|(X_1=x_1)$}
To sample $X_0$ we need to sample the number of elements, and the state of each element. While in general the state of $X_0$ elements can be coupled to $X_1$, here we only concretely explore schemes where each $X_0$ element is independently sampled from a base distribution, which is a `masked token' for all discrete state components, a standard isotropic Gaussian for all continuous components, and uniform samples on SO(3) for rotations.

To sample the $X_0$ length, we use:
$$\ell^{\text{group}}_0 \sim 1 + \text{Poisson}(\lambda_{\text{group}}),$$
where in many cases $\lambda_{\text{group}}$ is 0, meaning $X_0$ starts with a single element (or one per group).

\subsection{Deletions: $X_1^{+ø} \sim \mathcal{D}_{ø}|(X_1=x_1,X_0=x_0)$}

Sampling $X_1^{+ø} \sim \mathcal{D}_{ø}|(X_1=x_1,X_0=x_0)$ includes how many `to-be-deleted' elements are introduced, where they are introduced in the sequence, and what their state is, all which can be flexibly chosen. In what we think will provide a better learning target, we consider schemes where elements of $X_1$ are duplicated, and their state copied. In some cases it also might sense to modify the state, such as enforcing that `to-be-deleted' discrete states are always masked tokens.

Importantly, deletions are chosen such that $\ell^{\text{group}}_1 \ge \ell^{\text{group}}_0$ for all groups. To ensure this, one scheme we use is to specify a deletion rate $d_r \ge 1$, and draw the number of to-be-deleted elements with
$$\text{del}_{\text{group}} = \text{Poisson}(\max(\ell^{\text{group}}_0,\ell^{\text{group}}_1) d_r  -\ell^{\text{group}}_1)$$
$\text{del}_{\text{group}}$ elements are then sampled, uniformly with replacement, from the elements from the matching group of $X_1$, and inserted, uniformly, either side of $X_1$ template element, inheriting its state.

Another scheme we use, only when $\ell^{\text{group}}_0 = 1$, is to take each element in $X_1$ and duplicate it, inserting that randomly on either side.

\subsection{Forest: $\mathcal{T} \sim \mathcal{D}_{\mathcal{T}}|(X_1^{+ø}=x_1^{+ø},X_0=x_0)$}
We need $\ell^{\text{group}}_0$ trees in $\mathcal{T}$ (i.e. one tree per $X_0$ element). Since the trees must be planar, we adopt a backward coalescence scheme that starts with the elements of $X_1^{+ø}$ and uniformly selects adjacent pairs that belong to the same group, coalescing them and replacing the pair with a single node. This proceeds until there are only $\ell_0$ remaining elements, which become the root of each tree that is associated, maintaining the order, with the elements of $X_0$.

\subsection{Internal Anchors: $\mathcal{A^I} \sim \mathcal{D}|(X_1^{+ø}=x_1^{+ø},X_0=x_0, \mathcal{T})$}

Anchors control the conditional process along branches, guiding each element so that it would terminate, were $t$ to reach 1, at the anchor. Along terminal branches, the anchors are equal to the elements of $X_1^{+ø}$, but along internal branches we have freedom in how we construct them.

Consider a path along one tree from the root to a single leaf. An element drifts, along this path, towards one anchor after another, changing each time a split event occurs. If the anchors change too erratically, the process will be difficult to learn. One principle is to sample anchors that avoid, as much as possible, high-velocity changes.

Since anchors can be sampled conditioned on $X_1^{+ø}$, one convenient approach is to specify an anchor merging distribution $p(a_{\text{parent}}|a_{\text{child}_1},a_{\text{child}_2},\mathcal{T} )$ and recursively determine the anchors from leaves to root. This is depicted as $m(\cdot,\cdot)$ in figure \ref{fig:construction}.
Since ordered sequences of continuous values (especially for spatial positions) often exhibit strong autocorrelation with similar values for neighboring elements but discrete values typically do not, the anchor merging schemes can differ depending on whether the state components are continuous or discrete. For our empirical investigations, we use:

\begin{itemize}
    \item Continuous (\(\R^n\)): mean of $a_{\text{child}_1}$ and $a_{\text{child}_2}$, weighted by the number of descendants of each child.
    \item Manifold: Geodesic interpolation between $a_{\text{child}_1}$ and $a_{\text{child}_2}$, weighted by the number of descendants of each child.
    \item Discrete: All internal anchors are set as a mask/dummy token.
\end{itemize}

An alternative choice which we have explored, but not used in our examples below, is copying a random child (with probability optionally weighted by some function of the number of descendants of each child).

%% file: refactored_appendices/B_splits_deletions.tex
\section{Counting and Deletion Process Flow Matching}
\label{app:cdflow}

We first describe Counting Flows and Deletion Flows, which we will use to construct Branching Flows. 

\subsection{Counting Flows}\label{app:countingflow}
\subsubsection{Conditional distributions}
We first introduce `Counting Flows': flow matching over a Poisson counting process that generates samples from the space of non-negative integer `counts' via a conditional process that repeatedly increments by 1 to terminate at an observed count $z_{c}$ by $t=1$. This will later be repurposed for Branching Flows to model the counts of "splitting events" ahead of a node at time \(t\), but we present it here in isolation.

Let \(H_\mathrm{inc}\) be a hazard distribution with no atom at $1$ supported on \([0,1]\), implying that $S_{H_\mathrm{inc}}(1)= 1-F_{H_\mathrm{inc}}(1) = 0$. Since $S_{H_{\mathrm{inc}}}(t)$ is the probability of the event not occurring by time $t$, this implies the event occurs almost surely by $t=1$. The hazard rate associated to $H_\mathrm{inc}$ is given by \[h_\mathrm{inc}(t) = \frac{f_{H_\mathrm{inc}}(t)}{1-F_{H_\mathrm{inc}}(t)}.\]
We also note that since 
\[\int_0^t h(s) ds = -\log (1-F_{H_{\mathrm{inc}}}(t)),\]
a hazard $h(t) \geq 0$ gives rise to a hazard distribution supported on $[0,1]$ with no atom at $1$ as long as 
\[\int_0^1 h(t) dt = \infty. \]
Conditioned on an observed count \(z_{c}\) we construct a time-inhomogeneous CTMC on \(\{0,1,\dots,z_{c}\}\) with rates
\begin{align*}
    Q_t^{z_{c}}(x_t+1|x_t) &= (z_c-x_t)h_\mathrm{inc}(t) I_{\{x_t < z_c\}},
\end{align*}
and $Q_t^{z_c}(x_t|x_t) = -Q_t^{z_c}(x_t+1|x_t),$ with $Q_t^{z_c}(\cdot | x_t) = 0$ otherwise. 

For any $t$, the increment rate is thus proportional to the number of increments remaining, and the process will almost surely terminate at $z_{c}$.

With $R^{z_{c}}_t = z_{c}-x_t$ the remaining increments, and $R^\theta_t$ a neural network's prediction of the remaining increments, then

\begin{equation}\label{loss:counting}
    D_R(R_t^{z_c}, R_t^\theta) = R^\theta_t-R^{z_{c}}_t\log R^\theta_t
\end{equation}
is a valid loss under an appropriately scaled time-dependent linear parametrization \citep{billera2025timedependentlossreweighting}.

It follows that sampling single-increment independent and identically distributed unordered event times from \(H_\mathrm{inc}\) corresponds to sampling from the above CTMC, and for a given $t,x_t,z_{c}$ we can efficiently sample the waiting time until the next increment event.

\begin{alemma}
Let \(T_1, T_2, \dots, T_n \sim H\) be independent samples from the hazard distribution \(H\), with cumulative distribution function $F_H(t)$ and hazard $h(t)$. Define
\begin{equation*}
    X_t := \sum_{i=1}^{n} I_{\{T_i \leq t\}}.
\end{equation*}
Conditioned on $X_t = x_t$, there are exactly \(n-x_t\) remaining events. Denote $R:= n-x_t$. It holds that
\begin{equation*}
    \Prob[X_{t+\delta}=x_t+1|X_t=x_t] = Rh(t)\delta + o(\delta).
\end{equation*}  

\end{alemma}
\begin{proof}
    Let \(T_t^+ := \min\{T_i : T_i > t\}\) denote the arrival time of the next jump at time \(t\).   Condition on the specific at risk index set $A = \{i: 1\leq i \leq n , \  T_i > t\}$, corresponding to $|A| = R$ remaining events. Then for any $i\in A $ we have
    \begin{align*}
    p_\delta := \mathbb P(t < T_i \leq t+ \delta \mid T_i > t) &= 1 - \frac{S(t+\delta)}{S(t)} \\
    &= 1- \frac{S(t) + \delta S'(t) + o(\delta)}{S(t)} = h(t) \delta + o(\delta),
    \end{align*}
    noting that $S'(t) = -h(t)S(t) $. By independence, the number of jumps in $(t,t+\delta]$ among the $R$ at-risk units is $\mathrm{Binomial}(R, p_\delta)$, and thus 
    \begin{align*}
    \mathbb P(X_{t+\delta} =x_t + 1\mid X_t = x_t) &= Rp_\delta (1-p_\delta)^{R-1}.
    \end{align*}
    Since $p_\delta = h(t) \delta + o(\delta) = O(\delta)$ 
    \[(1-p_\delta)^{R-1} = 1 - (R-1)p_\delta + O(p_\delta^2) = 1 + o(1),\]
    it follows
    \[\mathbb P(X_{t+\delta} = x_t + 1 \mid X_t = x_t) = R(h(t) \delta + o(\delta))(1+o(1)) = Rh(t) \delta + o(\delta).\]
\end{proof}
\begin{acorollary}
    Fix $z_c \in \mathbb N_{\geq 0}$ and define $X_t = \sum_{i=1}^{z_c} I_{\{ T_i \leq t\}}$. Then it holds that $X_t$ evolves as a CTMC with rate matrix 
    \begin{align*}
        Q_t^{z_{c}}(x_t+1|x_t) &= (z_c-x_t)h(t) I_{\{x_t < z_c\}},
    \end{align*}
    and $Q_t^{z_c}(x_t|x_t) = -Q_t^{z_c}(x_t+1|x_t),$ with $Q_t^{z_c}(\cdot | x_t) = 0$ otherwise. In other words, if the event times are distributed according to $H$, then counting the events that have occurred up until the current time out of $z_c$ total events corresponds to the counting flow process conditioned on $z_c$, starting at zero. 
\end{acorollary}

\subsubsection{Interarrival Times}
\begin{atheorem}\label{thm:waitingtime}     Fix $n \geq 1$, and let $T_1,\ldots, T_n \sim H$ where $H$ is a hazard distribution having cumulative distribution function $F_H(t)$ and survival function $S_H(t) = 1-F_H(t)$. Set 
\[X_t = \sum_{i=1}^n I_{\{T_i \leq t\}}\]
and let $T_t^+ :=\min \{T_i : T_i > t\}$. Let $ V_t = (T_t^+ |X_t = x_t)$, and define $W_t = V_t - t$ as the conditional inter-arrival time for the next arrival, conditioned on $x_t$ preceding arrivals. 

Denote $R := n- x_t$ for the remaining number of events, given that $x_t$ events already happened and that there are $n$ events total. Then for each $t$, the quantile function of $V_t$ is given by 
\begin{align*}
F_{V_t}^{-1}(u) &=F_H^{-1}(1-S_H(t)(1-u)^\frac{1}{R}),
\end{align*}
where $F_H^{-1}(u) := \inf\{y : F_H(y) \geq u\}$ is the quantile function for $H$. In particular, if $U \sim \mathrm{Unif}(0,1)$, then 
\[W_t \overset d= F_H^{-1}\left(1- S_H(t)(1-U)^{\frac{1}{R}}\right) -t. \]
\end{atheorem}

\begin{proof}
Let $T_1,\ldots, T_R \sim H$ i.i.d. denote the $R := n - x_t$ arrival times exceeding $t$ and fix $t \in (0,1)$. For $y\geq t$, we have 
\[G_t(y) := \mathbb P(T_i \leq y \mid T_i > t) = \frac{F_H(y) - F_H(t)}{1-F_H(t) } = \frac{S_H(t) - S_H(y)}{S_H(t)}.\]
By independence,
\begin{align*}
\mathbb P(V_t \geq y) = \mathbb P( T_i \geq y\mid T_i > t)^R  =  \left(\frac{S_H(y)}{S_H(t)}\right)^R.
\end{align*}
The quantile function is $F_{V_t}^{-1}(u) = \inf\{y: F_{V_t}(y) \geq u\}$. It holds that 
\[F_{V_t}(y)  = u \iff 1- \left(\frac{S_H(y)}{S_H(t)}\right)^R = u  \iff F_H(y) =1- S_H(t)(1-u)^\frac{1}{R},\]
which allows us to conclude  
\[F_{V_t}^{-1}(u) =F_H^{-1}(1-S_H(t)(1-u)^\frac{1}{R}). \]
Let $U \sim \mathrm{Unif}(0,1)$. It is well known that passing $U$ to the quantile function recovers the original distribution: \[V_t \overset{d}{=} F_{V_t}^{-1}(U)\] so since $W_t = V_t - t$, we have
\[W_t \overset d= F_H^{-1}\left(1-S_H(t)(1-U)^\frac{1}{R}\right) - t.\]
\end{proof}
\begin{acorollary}\label{app-corr:efficientsampling}
    Fix $z_c \geq 1$ and let $X_t$ be a counting process with hazard distribution $H$. Condition on $X_t = x_t$, where $x_t < z_c$. Then there are $z_c - x_t$ remaining events and the waiting time until the next event has distribution
    \[W_t \overset d= F_H^{-1}\left(1-S_H(t) (1-U)^\frac{1}{z_c - x_t}\right) - t\]    where $U \sim \mathrm{Unif}(0,1)$. This allows us to efficiently sample the next waiting time given $t$, $x_t$ and $z_c$.
\end{acorollary}

\subsubsection{Marginal Generator}\label{app:margctmcgen}
Let $Z_c \sim p_{Z_c}(dz_c)$ and $Q_t^{z_c}$ specify a conditional on \(Z_c=z_c\) rate matrix of the counting flow process, \[Q_t^{z_c}(x_t +1 |x_t) = h(t)(z_c - x_t) I_{\{x_t < z_c\}},\]
and $Q_t^{z_c}(x_t|x_t) = - Q_t^{z_c}(x_t+1|x_t)$, with $Q_t^{z_c}(\cdot | x_t) = 0$ otherwise. We will recast this in terms of a time-dependent jump kernel (see Appendix~\ref{app:jump-measure}). This specifies a conditional process $X_t | Z_c=z_c$.

The time-dependent jump kernel characterizing the counting process is  
\[Q_t^{z_c}(dy ; x_t) = h(t)(z_c - x_t) \delta_{x_t +1 }(dy),\]
corresponding to jumping from $x_t$ to $x_t+1$ with rate $h(t)(z_c - x_t)$. It holds that all remaining events will occur by $t=1$ almost surely. Using the notion of a time-dependent linear parametrization \citep{billera2025timedependentlossreweighting}, we can parametrize the infinitesimal generator corresponding to the jump process above by $R_t^{z_c}(x) = z_c- x$, and have a valid conditional generator matching loss be given by
\[L_{\mathrm{cgm}}(\theta) = \mathbb E_{Z_c\sim p_{Z_c}, X_t \sim p_{t|Z_c}(dx| z_c)}[D(R_t^{Z_c}(X_t), R_t^\theta(X_t))],\]
where $R_t^\theta(x_t) $ is emitted by the model for each realization \(x_t\) of \(X_t\), and $D$ is a Bregman divergence. Note that the loss is minimized by the parameterization of the marginal generator, which coincides  by linearity with the posterior averaged parameterization of the conditional generator
\[R_t(x_t) = \mathbb E_{Z_c\sim p_{z_c}(dz_c| x_t)}[R_t^{Z_c}(x_t)] = \mu_{z_c|x_t} - x_t,\]
setting $\mu_{z_c|x_t} := \mathbb E_{Z_c\sim p_{z_c}(dz_c| x_t)} [Z_c] $.
The parameterization emitted by the model $F_t^\theta(x_t)$ corresponds to a time-dependent jump kernel 
\[Q_t(dy|x_t) = h(t) R_t^\theta(x_t)  \delta_{x_t+1}(dy).\]
The sampling procedure corresponding to the model-emitted jump kernel is
\[
X_0 \sim p_0, \qquad (X_{t+\Delta t}|X_t=x_t) = \begin{cases}
    x_t & \text{ with probability } 1 - \Delta t \cdot h(t) R_t^\theta(x_t), \\
    x_t +1 &\text{ with probability } \Delta t \cdot h(t)R_t^\theta(x_t),
\end{cases} 
\]
and when $F_t^\theta = F_t$ we have $X_t \sim p_t(dx)$ as $\Delta t \to 0$.

\subsection{Deletion Flows}
\label{subs:delflows}

We also describe a CTMC in the state space $S = \{0,1\}$, where 0 is `present' and 1 is `deleted'. Let $H_\mathrm{del}$ be a hazard distribution supported on $[0,1]$ with no atom at $1$, with its associated hazard rate \[h_\mathrm{del}(t)= \frac{f_{H_\mathrm{del}}(t)}{1-F_{H_\mathrm{del}}(t)}\] We represent the conditional probability of a deletion by $t=1$ with $Z_d \in \{0,1\}$. This determines the conditional rate $Q_t^{Z_d}(x|x_t)$, defined by:
\[Q^{Z_d}_t(1|0)  = Z_dh_\mathrm{del}(t)  \quad \text{and} \quad Q_t^{Z_d}(0|0) = - Q^{Z_d}_t(1|0),\] 
along with $Q^{Z_d}_t(\cdot |x_t) = 0$ for all other values of $x_t$. Conditional on $Z_d =1$, corresponding to a deleted element, the process $X_t|(Z_d=1)$ will terminate in deletion by $t = 1$ with probability \(1\). Let $\rho_t^{z_d} = z_d$ and $\rho_t^\theta$ be a neural network's prediction of the probability of deletion by time $t$. Then

\begin{equation*}
D_\rho(\rho_t^{z_d}, \rho_t^\theta) = -[z_d \log(\rho_t^\theta) + (1-z_d)\log(1-\rho_t^{\theta})]
\end{equation*}
is a valid, per-term, conditional generator matching loss under an appropriately scaled time-dependent linear parameterization \citep{billera2025timedependentlossreweighting}.

\begin{aremark}
    \label{remark:deltimes}
    The generator and the interarrival times obtained from having one or zero remaining events for counting flows will coincide for deletion flows.
\end{aremark}
\subsubsection{Branching Flows}

We construct Branching Flows with split and deletion events. The hazard rate of split events is defined identically to that of the increments in Counting Flows. For splits, the waiting time until the next event will coincide with that in Theorem~\ref{thm:waitingtime}, where the number of remaining split events ahead of each node corresponds to the number of remaining increments in Counting Flows. Deletions are then exactly as in Deletion Flows. 

%% file: refactored_appendices/C_branch_tracking.tex
\section{Branch tracking, splits, and deletions}
\label{app:branch-tracking}
We identify a branch with the node `ahead' of it (in the root-to-leaf, $t=0$ to $t=1$ direction), and $g_t^{(i)} = (\tau_t^{(i)}, b_t^{(i)})$ stores a tree index $\tau_t^{(i)}$, and a branch indicator vector $b_t^{(i)}$ comprising a possibly-empty ordered sequence of child indices $\uparrow $ (first child) or $\downarrow$ (second child) which, since the tree is binary, read from root towards leaves identifies a single branch on the tree by identifying which of the two children are descended at each bifurcation. An element of the augmented state reads $\tilde x_t^{(i)} = (x_t^{(i)},g_t^{(i)}) = (x_t^{(i)},(\tau_t^{(i)}, b_t^{(i)}))$.
With $<>$ the empty branch indicator tuple (using non-standard tuple brackets to make the tuple nesting easier to read), $(\tau_t^{(i)}=k, b_t^{(i)} = <>)$ denotes the root of the $k^{th}$ tree, and $(\tau_t^{(i)}=k, b_t^{(i)} = <\uparrow \uparrow \downarrow>)$ would denote the second child of the first child of the first child of the root of the $k^{th}$ tree.

Let $\tilde x_t^{(i)} = (x_t^{(i)}, (\tau_t^{(i)}, b_t^{(i)})).$ We use $\triangleright$ to denote the append operation\[\tilde x_t^{(i)} \triangleright \uparrow \ := (x_t^{(i)}, (\tau_t^{(i)}, b_t^{(i)}\triangleright \uparrow)), \]
and analogously define $\tilde x_t^{(i)}\triangleright\downarrow$.

For a state $\tilde x_t = (\tilde x_t^{(1)}, \ldots, \tilde x_t^{(n)})$, we define a `split operator':
\[\operatorname{split}_i(\tilde x_t ) := (\tilde x_t^{(1)}, \ldots, (\tilde x_t^{(i)}\triangleright \uparrow), \ (\tilde x_t ^{(i)}\triangleright\downarrow), \ldots, \tilde x_t^{(n)})).\]
This duplicates, in place, the element $\tilde x_t^{(i)}$, sending a copy down each of the two child branches. We also define a `deletion' operator: 
\[\operatorname{del}_i (\tilde x_t) := (\tilde x_t^{(1)}, \ldots, \tilde x_t^{(i-1)}, \tilde x_t^{(i+1)}, \ldots, \tilde x_t^{(n)}).\]
that deletes the element $\tilde x^{(i)}$ at index $i$.

%% file: refactored_appendices/D_base_processes.tex
\section{Base processes}
\label{app:baseproc}
\subsection{The Ornstein-Uhlenbeck process with Time Dependent Diffusion Coefficient}
\label{app:time-inhomo-ou}
Consider the stochastic differential equation

\begin{equation}\label{app-eqn:timeinhomosde}
    dX_t = \theta(\mu - X_t)dt + \sqrt{v_t}\,dW_t,
\end{equation}
where $\theta >0$ and \(v_t\) is the time-dependent infinitesimal variance (so the diffusion coefficient is \(\sqrt{v_t}\)). We will, for some \(x_1\), consider the time-inhomogeneous OU-bridge \(X_t\mid X_1=x_1\) where the unconditional process \(X_t\) solves
\begin{equation}\label{app-eqn:thesdeweactuallycareabt}
    dX_t = \theta(x_1 - X_t)dt +\sqrt{v_t}dW_t.
\end{equation}
A slightly more general problem is studied in \cite{AlbanoGiorno2020OU}, specifically of the form (See their equation (2.4)) 
\begin{equation}\label{app-eqn:albanogiornosde}
    dX_t = \left(-\frac{X_t}{\nu} + \mu + m_t\right)dt + \sqrt{v_t}dW_t,
\end{equation}
where explicit formulas for the transition densities (which remain Gaussian) are given, which we use to derive bridge densities. 
\subsection{Conditional and Marginal Paths for a Noisy Discrete Process}
\label{app:dfm}
In what follows, we use an extension of Discrete Flow Matching (DFM \cite{Gat2024DiscreteFM}) to allow time intervals $[t_0,t_1]$. Let $F_1$ and $F_2$ be CDFs of some continuous distributions supported on $[0,1]$, and let $\omega_u \in [0,1)$. Then we define the schedulers 
\[\kappa_1(t) = F_1(t), \qquad \kappa_2(t) = \omega_u(1-F_1(t))F_2(t), \qquad \kappa_3(t) = 1 - \kappa_1(t) - \kappa_2(t). \]
 It can be shown that our choice of schedulers enforces $\ell_\kappa := \mathrm{argmin}_{j\in [m]}\frac{\dot\kappa_t^j}{\kappa_t^j}  = 3$. With these schedulers, define the interpolant path 
    \[p_t(x|x_{t_0},x_1) := (\kappa_t^1-\kappa_{t_0}^1\frac{\kappa^3_t}{\kappa^3_{t_0}})\delta_{x_1}(x) + (\kappa_t^2-\kappa_{t_0}^2\frac{\kappa^3_t}{\kappa^3_{t_0}})p_u(x) + \frac{\kappa^3_t}{\kappa^3_{t_0}}\delta_{x_{t_0}}(x)\]
    between $x_{t_0}$ and $x_1$ in the time-interval $[t_0,1]$. This path is generated by 
    \[
    u_t(x,z|x_{t_0}, x_{t_1}) = (\dot \kappa_t^1 - \kappa_t^1\frac{\dot \kappa_t^3}{\kappa_t^3}) \delta_{x_{t_1}}(x) + (\dot \kappa_t^2 - \kappa_t^2\frac{\dot \kappa_t^3}{\kappa_t^3}) p_u(x) + \frac{\dot \kappa_t^3}{\kappa_t^3}\delta_{z}(x)
    \]
    which follows from applying DFM Theorem 3 on the interval $[t_0,1]$. To see this, define $\alpha_1 = \kappa_t^1 - \kappa_{t_0}^1\frac{\kappa_t^3}{\kappa_{t_0}^3}$, $\alpha_2 = \kappa_t^2 - \kappa_{t_0}^2 \frac{\kappa_t^3}{\kappa_{t_0}^3}$ and $\alpha_3 = \frac{\kappa_t^3}{\kappa^3_{t_0}}$. It can be seen that $\ell_\alpha := \mathrm{argmin}_{j\in [m]}\frac{\dot\alpha_t^j}{\alpha_t^j}  = 3 $, along with $\alpha_1,\alpha_2,\alpha_3\geq 0$ and $\sum_{i=1}^3 \alpha_i= 1$. The expression for the rates follows from
    \begin{align*}
            \dot \alpha_1 - \alpha_1\frac{\dot\alpha_3}{\alpha_3} &= (\dot\kappa_t^1 - \kappa_{t_0}^1 \frac{\dot\kappa_t^3}{\kappa_{t_0}^3}) - (\kappa_t^1 - \kappa_{t_0}^1\frac{\kappa_t^3}{\kappa_{t_0}^3})\frac{\dot\kappa^3_t}{\kappa^3_{t}}\\
            &= \dot\kappa_t^1 - \kappa_{t}^1 \frac{\dot\kappa^3_t}{\kappa^3_t}
            \intertext{and analogously}
            \dot \alpha_2 - \alpha_2\frac{\dot\alpha_3}{\alpha_3} &= \dot\kappa_t^2 - \kappa_{t}^2 \frac{\dot\kappa^3_t}{\kappa^3_t}.
    \end{align*}

%% file: refactored_appendices/E_generators.tex
\section{Branching Flows in Generator Matching} 

\subsection{The Infinitesimal Generator of Jump Processes}
\label{app:jump-measure}
In line with Generator Matching (GM) \citep{Holderrieth2024GeneratorMG} and Flow Matching Guide (FMG) \citep{Lipman2024FlowMG}, we model jumps by a time-dependent kernel $Q_t(dy; x)$ that, for each state $x\in S$, assigns a positive measure on $S \setminus \{x\}$. Its total mass \[\lambda_t(x) = \int Q_t(dy;x)\]
gives the jump intensity or jump rate, i.e. the instantaneous hazard of leaving $x$, and we require $\lambda_t(x) < \infty$. When $\lambda_t(x) >0$, normalizing $Q_t$ yields the jump destination distribution
\[J_t(dy; x) = \frac{Q_t(dy;x)}{\lambda_t(x)},\]
which is a probability measure on $S\setminus \{x\}$. Now we see how specifying a jump kernel $Q_t(dy;x)$ describes a jump process. With the hazard/rate of $\lambda_t(x)$, we will jump to a point $y \sim J_t(dy;x)$. 

For $\lambda_t(x_t) = 0$ the process is constant, and otherwise the sampling procedure is as follows:
\[(X_{t+\Delta t}|X_t=x_t) = \begin{cases}
    X_t & \text{with probability } 1-\Delta t\lambda_t(x_t) + o(\Delta t), \\
    \sim J_t(dy ; x_t) & \text{with probability } \Delta t\lambda_t(x_t) + o(\Delta t).
\end{cases}\]
The infinitesimal generator $\mathcal L_t$ associated to a process with jump measure $Q_t$ becomes 
\[\mathcal L_tf(x) = \int (f(y) - f(x))Q_t(dy; x) \]
As in \citep{billera2025timedependentlossreweighting} we consider atomic time-dependent jump kernels: \[Q_t(dy; x) = \sum_{i=1}^n \lambda_{i}(t,x) \delta_{\Gamma_i(t,x)}(dy),\]
    specifying \emph{jump targets} $\Gamma_i(t,x)$ and \emph{jump rates} $\lambda_{i}(t,x)$. The total rate is 
    \[\lambda_{\mathrm{total}}(t,x) := \int Q_t(dy; x) = \sum_{i=1}^n \lambda_i(t,x)\]
    and the normalized jump distribution $J_t(dy; x)$ is 
    \[J_t(dy;x ) = \frac{1}{\lambda_\mathrm{total}(t,x)}\sum_{i=1}^n\lambda_i(t,x) \delta_{\Gamma_i(t,x)}(dy), \]
    so that if $Y \sim J_t(dy|x)$, then $\mathbb P(Y = \Gamma_i(t,x)) = \frac{\lambda_i(t,x)}{\lambda_{\mathrm{total}}(t,x)}$ for $i=1,\ldots, n$.

\subsection{The Marginal Branching Infinitesimal Generator Rewrite}\label{app:linparambranch}
Let $S := \bigsqcup_{n\geq 1} \mathcal E^n$. We say that $x\in \mathcal E^n$ has $n$ elements. We let $h_\mathrm{split}(t)$ and $h_\mathrm{del}(t)$ be overall hazard multipliers as in Appendix~\ref{app:countingflow}, corresponding to an overall rate multiplier for split increments and an overall rate multiplier for deletions. Suppose that splitting of the $i$'th element out of the $n$ elements occurs with rate $h_\mathrm{split}(t)\cdot R_{t,i}(x)$ and that deletion of the $i$'th element out of the $n$ elements occurs with rate $ h_\mathrm{del}(t) \cdot \rho_{t,i}(x)$. Here, $R_{t,i}(x) \geq 0$ is intended to indicate the `remaining number of splits of the $i$'th element' and $\rho_{t,i}(x) \in [0,1]$ is intended to indicate the `probability of deleting the $i$'th element'. For $n=1$, we require that the deletion probability is zero so as not to result in an empty sequence.  
In this section, we arrive at a time- and state varying linear parametrization for the infinitesimal generator of our branching process. The notion of a linear parametrization of a generator, introduced in \cite{Holderrieth2024GeneratorMG} and further discussed in \cite{Lipman2024FlowMG}, was clarified to be able to vary with both time and state in \cite{billera2025timedependentlossreweighting}. 

For the subsequent discussion, we consider times $t$ such that $h_\mathrm{split}(t) > 0$ and $h_{\mathrm{del}}(t) >0 $. This is justified because if the hazard rate for split events $h_\mathrm{split}(t)$ is zero, there is nothing for the model to learn since in that case it is known \emph{a priori} that the split rate $h_\mathrm{split}(t) \cdot R_{t,i}(x)$ vanishes. Similarly, if the hazard rate for deletion events $h_\mathrm{del}(t)$ is zero, the deletion rate $h_\mathrm{del}(t)\cdot \rho_{t,i}(x) =0$ is also known \emph{a priori}.  

In what follows, we consider $x\in  \mathcal E^n $ having $n\geq 1$ elements. A time- and state varying linear parametrization of the branching generator is
\begin{align*}
\mathcal L^\mathrm{branch}_tf(x) &= \langle \mathcal K_{t,n}^\mathrm{split}   f(x), R_t(x) \rangle_{t,n}^\mathrm{split} + \langle \mathcal K_{t,n}^\mathrm{del}f(x), \rho_t(x)\rangle_{t,n}^\mathrm{del} + \langle \mathcal K_{t,n}^\mathrm{base} f(x), F_t^\mathrm{base}(x)\rangle_{t,x}^\mathrm{base}.
\end{align*}
In the above, using an atomic time-dependent jump kernel as in \citep{billera2025timedependentlossreweighting}, splits and deletions correspond to the time-dependent jump kernel
\begin{align*} 
Q_{t}(dy;x) &=  \sum_{i=1}^{n} h_\mathrm{split}(t)R_{t,i}(x) \delta_{\mathrm{split}_i(x)}(dy) + \sum_{i=1}^n h_\mathrm{del}(t)\rho_{t,i}(x) \delta_{\mathrm{del}_i(x)}(dy).
\end{align*}
With this time-dependent jump kernel, the infinitesimal generator for jumps becomes
\begin{align*}
    \mathcal L_t^{\mathrm{jump}}f(x) 
    &= \int_S (f(y)-f(x)) Q_{t}(dy;x) \\
    &= \int_S \bigg((f(y) - f(x)) \bigg[h_\mathrm{split}(t)\cdot  \sum_{i=1}^{n}\Big( R_{t,i}(x) \delta_{\mathrm{split}_i(x)}(dy) \Big)\\
    &\qquad+ h_\mathrm{del}(t) \cdot \sum_{i=1}^n \Big( \rho_{t,i}(x) \delta_{\mathrm{del}_i(x)}(dy)\Big)\bigg]\bigg)  \\
    &= h_{\mathrm{split}}(t) \cdot \sum_{i=1}^{n} \Big((f(\mathrm{split}_i(x)) - f(x))R_{t,i}(x)\Big) \\
    &\qquad + h_{\mathrm{del}}(t) \cdot \sum_{i=1}^n \Big( (f(\mathrm{del}_i(x)) - f(x))\,\rho_{t,i}(x)\Big).
\end{align*}
To linearly parametrize this generator, we define the inner products \[\langle v,w \rangle_{t,n}^\mathrm{split}{} := h_\mathrm{split}(t) \langle v,w \rangle_{\mathbb R^n} \qquad\text{and}\qquad \langle v,w\rangle_{t,n}^\mathrm{del} := h_{\mathrm{del}}(t) \langle v,w\rangle_{\mathbb R^n}.\] 

This allows us to linearly parametrize the jump generator with
\[\mathcal L_t^\mathrm{jump}f(x) = \langle \mathcal K_{t,n}^\mathrm{split} f(x), R_t(x) \rangle_{t,n}^\mathrm{split} + \langle \mathcal K_{t,n}^\mathrm{del}f(x), \rho_t(x)\rangle_{t,n}^\mathrm{del},\]
where in the above, we set
\begin{align*}
    \mathcal K_{t,n}^\mathrm{split}f(x) &=(f(\mathrm{split}_i(x)) - f(x))_{i=1}^n
    \intertext{and}
    \mathcal K_{t,n}^\mathrm{del}f(x) &= (f(\mathrm{del}_i(x)) - f(x))_{i=1}^n.
\end{align*}
In between jumps, we assume that the process evolves according to a base generator on each branch level --- this could, for example, correspond to an SDE. We assume that we have a linear parametrization
\[\mathcal L_t^\mathrm{base}f(x) = \langle \mathcal K_{t,n}^\mathrm{base} f(x), F_t^\mathrm{base}(x)\rangle_{t,x}^\mathrm{base} \]
for the base process along each branch. This results in 
\begin{align*}
\mathcal L^\mathrm{branch}_tf(x) &= \mathcal L_t^\mathrm{jump}f(x) + \mathcal L_t^\mathrm{base}f(x)\\
&= \langle \mathcal K_{t,n}^\mathrm{split}   f(x), R_t(x) \rangle_{t,n}^\mathrm{split} + \langle \mathcal K_{t,n}^\mathrm{del}f(x), \rho_t(x)\rangle_{t,n}^\mathrm{del} + \langle \mathcal K_{t,n}^\mathrm{base} f(x), F_t^\mathrm{base}(x)\rangle_{t,x}^\mathrm{base} .
\end{align*}

\subsection{Conditioning on Latent Discrete Processes}\label{app:auggm}

In Branching Flows, the branching structure is encoded in the trees in $Z$, but the timing of the bifurcation events is stochastic. This means that, given the trees in $Z$ and a state $X_t$, each element in $X_t$ cannot be unambiguously assigned to a branch on a tree. We thus need to augment the state with a branch-tracking index $G_t$. While it is possible to additionally condition the model on the branch tracking index, this places an architectural burden on the model. To avoid this, we learn a generator on $X$ that marginalizes over the time-dependent auxiliary process $G_t$, extending the framework of Edit Flows~\citep{Havasi2025EditFF} from the purely discrete $X_t$ case to where $X_t$ can be any process allowed under Generator Matching, including drift, diffusion, and jumps.

With joint conditional paths $(X_t, G_t)|(Z=z) \sim  p_{t\mid Z}(\mathrm{d}x,\mathrm{d}g\mid z)$, we train a model that marginalizes over $G_t$ and $Z$, learning a marginal generator $\mathcal L_t$ that generates a process $\widehat{X}_t$ with marginals $p_t(\mathrm{d}x)$ that generates samples from the data distribution \( q(dx) \). In what follows, we will assume that the Generator Matching regularity conditions apply to the marginal process $\widehat{X}_t$ and the joint marginal $(X_t, G_t)$ (marginalizing over $Z$), and the joint conditional $(X_t, G_t)|(Z=z)$.

Let $\mathcal G$ be discrete with $|\mathcal G | <\infty$ and let $(X_t, G_t)$ be an $S\times \mathcal G$-valued process generated by $W_t: \mathcal T(S\times \mathcal G) \to C(S \times \mathcal G, \mathbb R)$. For fixed $g\in \mathcal G$, define $\mathcal L_t^gf(x):= W_t\tilde f(x,g) $, where $\tilde f(x,g) = f(x)$. 

In what follows, we define the measure $\mu^g_{X_t}(A) := p_{X_t,G_t}(A\times \{g\}) $. Note that for each $E \subset \mathcal G$ we have $p_{X_t,G_t}(dx,E) = \sum_{g\in E} \mu^g_{X_t}(dx) $. The following lemma holds: 
\begin{alemma}\label{lemma:KFEthing}
        \[\partial_t \sum_{g\in \mathcal G}\int_{S} f(x) \mu^g_{X_t}(dx) = \sum_{g\in \mathcal G}\int_{S} \mathcal L_t^{g} f(x) \mu^g_{X_t}(dx).\]
\end{alemma}
\begin{proof}Write 
\begin{align*} \partial_t \sum_{g\in \mathcal G}\int_{S} f(x) \mu^g_{X_t}(dx) &= \partial_t \int_{S} f(x) \sum_{g\in \mathcal G}\mu^g_{X_t}(dx) \\ \intertext{where in the above we are integrating with respect to the measure $\sum_{g\in \mathcal G}\mu_{X_t}^g(dx)$. This becomes} &= \partial_t \int_{S} f(x) p_t(dx,\mathcal G)\\ &= \partial_t \int_{S}f(x) \int_{\mathcal G}p_t(dx,dg)\\ &= \partial_t \int_{S\times \mathcal G}f(x) p_t(dx,dg) \\ &= \partial_t \int_{S\times \mathcal G} \tilde f(x,g) p_t(dx,dg)\\ 
\intertext{and since $W_t$ generates $p_t(dx,dg)$, by the Kolmogorov Forward Equation, it holds:} &= \int_{S\times \mathcal G} W_t\tilde f(x,g) p_{X_t,G_t}(dx,dg)\\ &= \sum_{g\in \mathcal G}\int_{S} \mathcal L_t^{g} f(x) \mu^g_{X_t}(dx). 
\end{align*} 
\end{proof}

\begin{atheorem}\label{thm:prop1}
    It holds that 
    \[\mathcal L_tf(x) := \sum_{g \in \mathcal G } \mathcal L_t^{g}f(x) \, p_{G_t|X_t}(g|x)\]
    generates $p_t(dx)$, i.e., $\mathcal L_t$ solves the KFE for $p_t$:
    \[\partial_t\int f(x)p_t(dx) = \int \mathcal L_tf(x) p_t(dx).\]
    Moreover, assume that $F_t^{g} : S\to \Omega \subset V $ is a linear parametrization of $\mathcal L_t^{g}f(x)$ over a vector space $V$. If there is a linear $\mathcal K : \mathcal T \to \mathcal C(S;V)$ on the inner product space $\langle \cdot,\cdot\rangle_V$ such that \[\mathcal L_t^{g}f(x) = \langle \mathcal Kf(x),  F_t^{g}(x)\rangle_{V}\]
    for each $g\in \mathcal G$, then \[F_t(x) = \mathbb E_{G_t\sim p_{G_t|X_t}(dg|x)} [F_t^{G_t}(x)]\] linearly parametrizes $\mathcal L_t$.                                                  
\end{atheorem}

\begin{proof}
    For each $t\in [0,1]$, we have $p_t(dx) =\sum_{g \in \mathcal G} \mu^g_{X_t}(dx)$, and hence:
    \begin{align*}
        \partial_t\int_S f(x) p_t(dx) &= \partial_t \int_S f(x) \sum_{g\in \mathcal G}\mu^g_{X_t}(dx)  \\
        &=\partial_t \sum_{g\in \mathcal G}  \int_S f(x)\mu^g_{X_t}(dx) \\
        \intertext{and by Lemma~\ref{lemma:KFEthing}, it holds:}
        &= \sum_{g\in \mathcal G} \int_S \mathcal L_t^{g}f(x) \mu^g_{X_t}(dx)\\
        &= \sum_{g\in \mathcal G}\int_S \mathcal L_t^{g}f(x) p_{G_t|X_t}(g|x) p_{t}(dx) \\
        &= \int_S \underbrace{\left(\sum_{g\in \mathcal G}\mathcal L_t^{g}f(x) p_{G_t|X_t}(g|x)\right)}_{=\mathcal L_tf(x)}p_t(dx).
    \end{align*}
    By the linearity of expectations and the map $w\mapsto \langle v,w\rangle_V$ for fixed $v\in V$, we have 
    \begin{align*}
    \mathcal L_tf(x) &= \mathbb E_{G_t\sim p_{G_t|X_t}(dg|x)}[\mathcal L_t^{G_t}f(x)] \\
    &=\mathbb E_{G_t\sim p_{G_t|X_t}(dg|x)}[\langle \mathcal Kf(x), F_t^{G_t}(x)\rangle_V] \\
    &= \langle \mathcal Kf(x), \mathbb E_{G_t\sim p_{G_t|X_t}(dg|x)}[F_t^{G_t}(x)]\rangle_V \\
    &= \langle \mathcal Kf(x), F_t(x)\rangle_V .
    \end{align*}
\end{proof}
\begin{aremark}
    As shown above, the process $\widehat X_t$ generated by $\mathcal L_t$ has marginals $p_t(dx)$, but we simply write our expectations over $X_t \sim p_t(dx)$ for clarity of notation.   
\end{aremark}
\begin{atheorem}\label{thm:prop2}
    The auxiliary-process-conditioned generator matching loss and the generator matching loss coincide up to a constant in $\theta$: 
    \[\nabla_\theta \mathbb E_{t\sim \mathcal D, X_t,G_t \sim p_t(dx,dg)}D(F_t^{G_t}(X_t), F_t^\theta(X_t)) = \nabla_\theta\mathbb E_{t\sim \mathcal D,  X_t\sim p_t(dx)} [D(F_t(X_t), F_t^\theta(X_t))].\]
\end{atheorem}

\begin{proof}
    Since expectations commute with the left slot of Bregman divergences under the gradient (cf. \cite{Lipman2024FlowMG}, \cite{billera2025timedependentlossreweighting}), we obtain:
    \begin{align*}
        \nabla_\theta \mathbb E_{t\sim \mathcal D, X_t,G_t \sim p_t(dx,dg)}D(F_t^{G_t}(X_t), F_t^\theta(X_t)) &= \nabla_\theta \mathbb E_{t\sim\mathcal D, X_t\sim p_t(dx)} \mathbb E_{G_t\sim p_{G_t|X_t}(dg|x)} D(F_t^{G_t}(X_t),F_t^\theta(X_t))\\ 
        &= \mathbb E_{t\sim\mathcal D, X_t\sim p_t(dx)} \nabla_\theta
        \mathbb E_{G_t\sim p_{G_t|X_t}(dg|x)} D(F_t^{G_t}(X_t), F_t^\theta(X_t))  \\ 
        &=   \mathbb E_{t\sim\mathcal D, X_t\sim p_t(dx)}\nabla_\theta D(\mathbb E_{G_t\sim p_{G_t|X_t}(dg|x)} [F_t^{G_t}(X_t)],F_t^\theta(X_t))\\
        &=  \mathbb E_{t\sim \mathcal D, X_t\sim p_t(dx)} \nabla_\theta D(F_t(X_t), F_t^\theta(X_t))\\
        &= \nabla_\theta\mathbb E_{t\sim \mathcal D, X_t\sim p_t(dx)} D(F_t(X_t), F_t^\theta(X_t)).
    \end{align*}
\end{proof}
\begin{aremark}
    The arguments in \cite{Holderrieth2024GeneratorMG} allow us to include an additional latent conditioning variable $z\in\mathcal Z$ that is often used to steer the path towards terminating at a particular point of the data distribution. Let $\mathcal L_t^{z,g}$ denote the conditional generator given $z\in \mathcal Z$ and $g\in \mathcal G$. Define the $z$-conditioned generator
    \[\mathcal L_t^{z}f(x) := \mathbb E_{G_t\sim p_{G_t|X_t,Z}(dg|x,z)}[\mathcal L_t^{z,G_t}f(x)].\]
    Then, by Theorem~\ref{thm:prop1} and Proposition 1 of GM,
    \begin{align*} 
    \mathbb E_{Z \sim p_{Z|X_t}(dz|x),G_t \sim p_{G_t|Z,X_t}(dg|z,x)}[\mathcal L_t^{Z,G_t}f(x)] &= \mathbb E_{Z\sim p_{Z|X_t}(dz|x)}[\mathbb E_{G_t \sim p_{G_t|Z,X_t}(dg|z,x)}[\mathcal L_t^{Z,G_t}f(x)]]\\ 
    &= \mathbb E_{Z\sim p_{Z|X_t}(dz|x)}[\mathcal L_t^{Z}f(x)] \\
    &= \mathcal L_tf(x),
    \end{align*}
    so averaging over a fixed $z\in \mathcal Z$ and a discrete process $(G_t)_{0\leq t\leq 1}$ recovers the same marginal generator $\mathcal L_t$. For the loss, if we write $F_t^{z,g}$ for a linear parametrization of $\mathcal L_t^{z,g}$, the same
    linearity argument as in Theorem~\ref{thm:prop2} gives
    \begin{align*}
    \nabla_\theta \mathbb E_{t\sim\mathcal D[0,1], X_t,Z,G_t} D(F_t^{Z,G_t}(X_t),F_t^\theta(X_t)) &=  \mathbb E_{t\sim\mathcal D, X_t} [\nabla_\theta\mathbb E_{Z \sim p_{Z|X_t}(dz|x),G_t \sim p_{G_t|Z,X_t}(dg|z,x)} [D(F_t^{Z,G_t}(X_t),F_t^\theta(X_t))]] \\
   &= \mathbb E_{t\sim\mathcal D, X_t} [ \nabla_\theta D(\mathbb E_{Z \sim p_{Z|X_t}(dz|x),G_t \sim p_{G_t|Z,X_t}(dg|z,x)}[F_t^{Z,G_t}(X_t)],F_t^\theta(X_t))] \\
    &= \mathbb E_{t\sim\mathcal D, X_t} \nabla_\theta D(F_t(X_t),F_t^\theta(X_t))
    \\
    &= \nabla_\theta \mathbb E_{t\sim\mathcal D, X_t} D(F_t(X_t),F_t^\theta(X_t)),
    \end{align*}
    so conditioning on a fixed $z\in\mathcal Z$ does not change the gradient of the loss. 
\end{aremark}

\subsection{The Branching Flows Loss}\label{app:branchingflowsloss}

The conditional branching flows loss is the auxiliary-process-conditioned generator matching loss associated to the linear parametrization from Appendix~\ref{app:linparambranch}, with conditional dynamics as described in Appendix~\ref{app:auggm}. For $x \in \bigsqcup_{n\geq 1} \mathcal E^n$, we specify a conditional infinitesimal generator of the form
\begin{align*}
\mathcal L^{z,g,\mathrm{branch}}_t f(x) &= \langle \mathcal K_{t,n}^\mathrm{split}   f(x), R_t^{z,g}(x) \rangle_{t,n}^\mathrm{split} + \langle \mathcal K_{t,n}^\mathrm{del}f(x), \rho_t^{z,g}(x)\rangle_{t,n}^\mathrm{del} + \langle \mathcal K_{t,n}^\mathrm{base} f(x), F_t^{z,g,\mathrm{base}}(x)\rangle_{t,x}^\mathrm{base}.
\end{align*}
and we similarly specify a model parametrized infinitesimal generator
\begin{align*}
\mathcal L^{\theta,\mathrm{branch}}_t f(x) &= \langle \mathcal K_{t,n}^\mathrm{split}   f(x), R_t^\theta(x) \rangle_{t,n}^\mathrm{split}  + \langle \mathcal K_{t,n}^\mathrm{del}f(x), \rho_t^\theta(x)\rangle_{t,n}^\mathrm{del} + \langle \mathcal K_{t,n}^\mathrm{base} f(x), F_t^{\theta,\mathrm{base}}(x)\rangle_{t,x}^\mathrm{base} .
\end{align*}
As in \citep{billera2025timedependentlossreweighting}, when a generator is expressed as a sum of linear parametrizations, a valid per-term conditional generator matching loss is then the sum of Bregman divergences
\begin{align*}
    D^{\mathrm{split}}_{t,x}(R_t^{z,g}(x), R_t^\theta(x)) + D^\mathrm{del}_{t,x}(\rho_t^{z,g}(x), \rho_t^\theta(x)) + D^\mathrm{base}_{t,x}(F_t^{z,g,\mathrm{base}}(x), F_t^{\theta,\mathrm{base}}(x)),
\end{align*}
again noting that $x$ stands for $x\in S$. Assuming that $R_t^{z,g}(x)$ can be interpreted as the remaining number of splits ahead of each element and that $\rho_t^{z,g}(x)$ can be interpreted as deletion probabilities, we choose a Poisson-like Bregman for splits and a cross-entropy loss for deletions.

Note that these loss functions require a notion of Bregman divergence associated with convex functions that are only differentiable on a subset of their domain, as considered in \citep{billera2025timedependentlossreweighting}

If we also assume that the base Bregman divergence is separable and splits along each ahead, the conditional branching flows loss becomes:
\begin{align*}
L_{\mathrm{CBF}}(\theta)  &= \mathbb E_{
t\sim \mathcal{D}[0,1],
Z\sim p_Z,
 (X_t,G_t)\sim p_{X_t,G_t|Z}(dx,dg|z)
}\\
&\qquad\Big[\sum_{i=1}^{L_t}\Big(
{R_{t,i}^\theta( X_t)-R_{t,i}^{Z,G_t}(X_t)\log R_{t,i}^\theta(X_t)}\Big)\\
&\qquad+\sum_{i=1}^{L_t} \Big(- [\rho^{Z,G_t}_{t,i}(X_t)\log \rho_{t,i}^\theta(X_t) + (1-\rho^{Z,G_t}_{t,i}(X_t))\log(1-\rho_{t,i}^\theta(X_t))]\Big) \\
&\qquad+ \sum_{i=1}^{L_t} \Big(D_{t,X_t, i}^{\mathrm{base}}(F_{t,i}^{Z,G_t,\mathrm{base}}(X_t),F_{t,i}^{\theta,\mathrm{base}}( X_t))
\Big)\Big].
\end{align*}

%% file: refactored_appendices/F2_analyses.tex
\begingroup
\makeatletter
\let\0\@undefined
\let\1\@undefined
\let\2\@undefined
\let\3\@undefined
\let\4\@undefined
\makeatother

\section{Supplementary Empirical Details}
\subsection{QM9 analysis} \label{dat:qm9}

\subsubsection{Data}
We use QM9 data (license: CC0) \citep{qm9pack}, where the atom order in the coordinate files is arranged to match the canonical SMILES string ordering, where hydrogens are after the heavy atoms. We reasoned that the splitting mechanism of Branching Flows would not excel if there are spatially distant elements that are neighbors in the primary sequence, since a late split would then require very large flow velocities. We thus kept the primary heavy atom ordering (since neighboring SMILES atoms are often bonded, and thus spatially close), but redistributed the hydrogens such that each hydrogen is inserted into the sequence in front of its nearest heavy atom, ordered by distance when there are more than one hydrogen neighbour per heavy atom.

\subsubsection{Branching Flows specification}
For QM9, we constructed our `default' Branching Flow with the following choices:
\begin{outline}
 \1 Conditional base process:
    \2 Continuous: OU process flow (Appendix \ref{app:time-inhomo-ou}), with a mean-reversion rate $\theta=5$, and where the variance decays from 10 when $t=0$ to $0.001$ by $t=1$.
    \2 Discrete: DFM convex interpolation (appendix \ref{app:dfm}) of $x_0$, uniform noise, and $x_1$, with hazards $\mathcal{D}_1=\text{Beta}(2,2), \mathcal{D}_2=\text{Beta}(2,2), \omega_u=0.2$.
\1 Branching hazard distribution: $\operatorname{Beta}(1,3/2)$
\1 Deletion hazard distribution: $U(0,1)$
\1 To-be-deleted policy: Poisson with mean 20\% of the $x_1$ length elements are sampled (with replacement) and duplicated, and the duplicated elements are inserted either side of the original.
\1 $x_0$ state distribution: $N(0,1)$ for each continuous coordinate, and mask/dummy token for the discrete atom label. 
\1 $x_0$ length distribution: 1, always.

\1 Anchors: Weighted average for continuous states, and masked token for discrete states.
\end{outline}

\subsubsection{Model and Training}\label{qm9-app:qm9modelandtraining}
We use a 12 layer transformer model (12 heads, each with head dimension of 64), with an embedding dimension of 384. The atom positions are encoded by Random Fourier Features (RFFs), and pairwise spatial features are also provided to each layer by a learnable, layer-specific, head-specific attention bias. Rotary Positional Encoding (RoPE, \cite{su2023roformerenhancedtransformerrotary}) is used to encode the primary sequence ordering. The final six layers additively update the atom positions (and the spatial pair features are recalculated after each position update). The model outputs the endpoint-predicted (i.e. by $t=1$) locations, logits over the endpoint-predicted discrete state, the log of the per-element expected number of future bifurcations by $t=1$, and a deletion binary logit. The model was trained, using Muon \citep{jordan2024muon} with post-warmup learning rate of $0.005$, for 500k batches (where each batch was 128 randomly-sampled QM9 molecules) and a linear learning-rate cooldown for the last 50k batches. For sampling, 10k samples from the Branching Flows QM9 model were generated with a time step schedule $t = 1-(\cos(\pi s)+1)/2$ for $s \in (0, 1/1000, 2/1000, \ldots,  1)$.

We compare generated samples to the data distributions of atom counts (total, carbon, nitrogen, oxygen, fluorine, and hydrogen), and molecular properties. For molecule properties, since the training samples and model-generated samples are atom point clouds (ie. no bond structure is explicitly represented), we use OpenBabel to convert .xyz coordinates to .sdf, which also infers bond structure and SMILES strings. The .sdf is then read into RDKit for downstream analysis, and computing molecule properties, and fingerprints. OpenBabel was run using the OpenBabel.jl Julia wrapper, and RDKit was run using the MoleculeFlow.jl Julia wrapper.

Molecular fingerprints encode the structural features of a molecule into a fixed-length bit vector which can be used for similarity search, clustering, etc. Here, we compute the `RDKit fingerprints` (using MoleculeFlow.jl) and embed these in two dimensions, using Uniform Manifold Approximation (run via UMAP.jl, with default settings). This is used to visualize the distribution of true QM9 and generated samples.

\begin{figure}
    \centering{\includegraphics[width=0.99\textwidth]{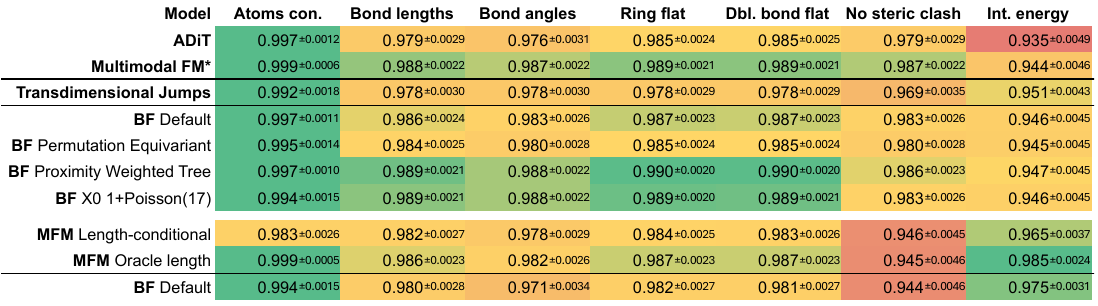}}
    \caption{\textbf{Posebusters metrics} from the models in Figure \ref{fig:QM9maintable}. Uncertainty is shown with symmetric two-sigma Wald intervals.}
    \label{fig:posebusters}
\end{figure}

We compute a multivariate Jensen–Shannon divergence (JSDm) similar to \citep{Schneuing2025MultidomainDL} between the generated and reference QM9 distributions using the filtered subset of molecules only. The calculation uses the joint distribution of 10 properties: molecular weight, logP,
  hydrogen-bond donors, hydrogen-bond acceptors, the numbers of C/H/O/N/F atoms, and the total atom count. For each property, bin boundaries are defined by reference-data quantiles (4 bins per property), and both datasets are then mapped into
  the resulting 10-dimensional joint histogram. JSDm is the Jensen–Shannon divergence between these two joint histograms, reported in bits; lower values indicate better agreement with the QM9 reference distribution.

Following ADiT \citep{joshi2025allatomdiffusiontransformersunified} we also include Posebusters \citep{Buttenschoen_2024} metrics, and Validity and Uniqueness. For each sample that can be converted to SDF, we run PoseBusters and record the individual pass/fail outcomes of the standard molecular geometry and chemistry checks. Validity is the fraction of all generated samples whose reconstructed molecule yields a valid SMILES string under the analysis pipeline. This is also normalized by the full sample count. Uniqueness is the fraction of distinct SMILES strings among the SMILES-valid samples only.

\begin{figure}[t!]
\centering{\includegraphics[width=0.99\textwidth]{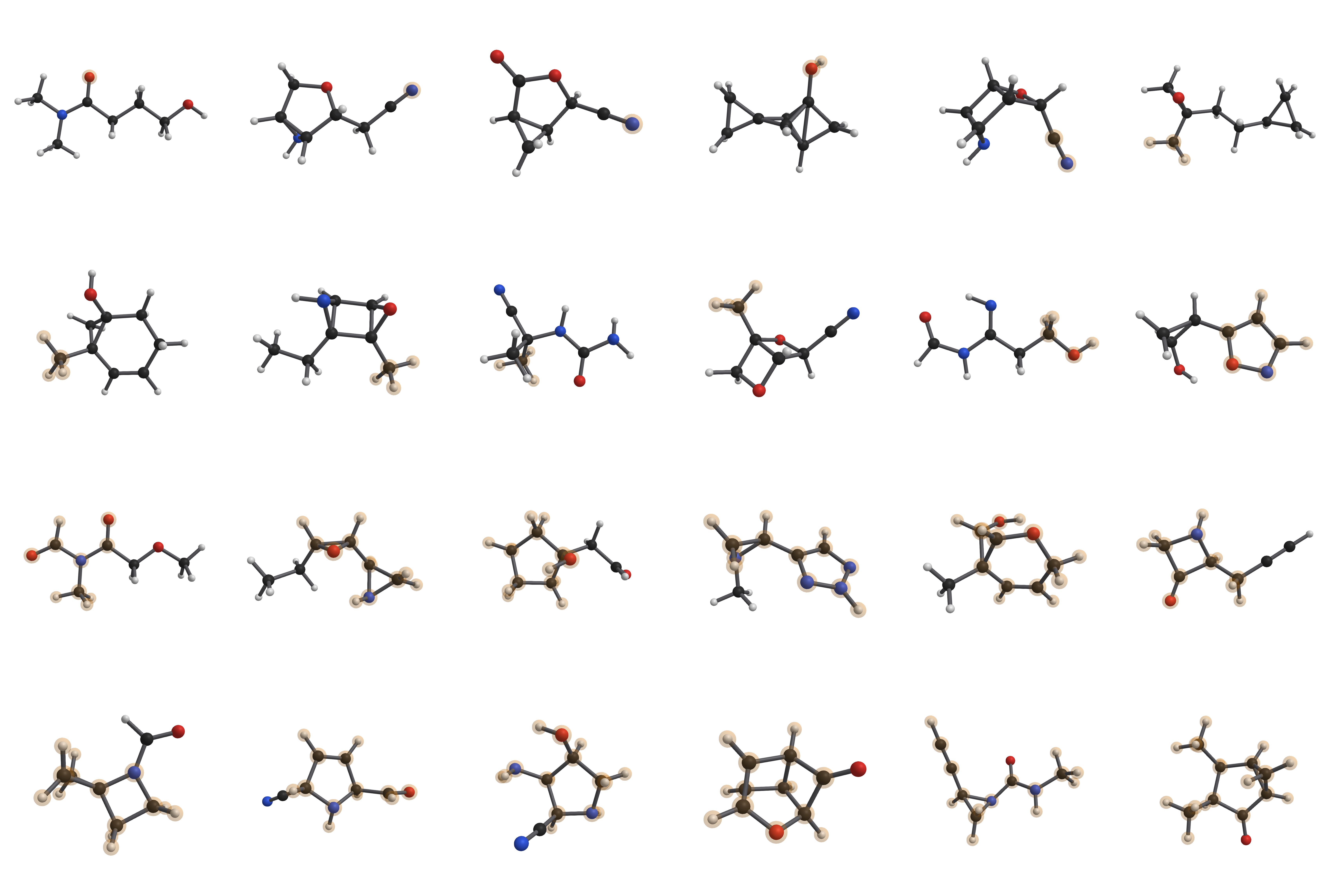}}
    \caption{\textbf{Examples of `Completion' task samples.} Spheres indicate designable atoms.}
    \label{fig:completion_examples}
\end{figure}

To quantify distribution overlap, we use a Kolmogorov-Smirnov-like statistic between the empirical cumulative distribution functions of samples $Y$ from each model against QM9 data. We use $1-KS_D$, where
$$KS_D = \max_{x}|\hat{F}_Y(x) - \hat{F}_{QM9}(x)|$$
where \(\hat{F}\) denotes the empirical CDF. This statistic is 1 when the distributions agree perfectly, and 0 when they do not overlap at all.

For Bernoulli metrics the tables report symmetric two-sigma Wald intervals:
  \[
  \hat{p} \pm 2\sqrt{\frac{\hat{p}(1-\hat{p})}{N}}.
  \]

We use 10,000 samples from ADiT \citep{joshi2025allatomdiffusiontransformersunified} obtainable directly from their code page.
Additionally, we used the model from \cite{Campbell2023TransDimensionalGM} to generate 10,000 samples, and analyze them as above.

\paragraph{QM9 Completion task:}
The QM9 completion experiments use the same basic generative setup as the unconditional baseline, but with an additional binary conditioning mask indicating which atoms are fixed and which atoms are designable. During training, fixed atoms
  are provided as context, while the model is trained to generate only the designable portion of the molecule. The conditioning masks are derived from a precomputed library of chemically sensible partial structures for each QM9 molecule. To build this library, each QM9 molecule is first reordered so that hydrogens are placed adjacent to their nearest heavy atoms, yielding contiguous heavy-atom blocks. Candidate masked regions are then defined as contiguous runs of these blocks whose heavy atoms form a connected subgraph and attach to the rest of the molecule through exactly one bond. This construction biases the candidates toward substituent-like pendant fragments rather than arbitrary atom subsets. 

  At training time, one first samples a molecule that has at least one valid candidate region. Then one candidate region is selected uniformly from that molecule’s candidate list. Finally, one of two conditioning polarities is chosen with equal
  probability: either the candidate interval is fixed and the rest of the molecule is designable, or the complement is fixed and the candidate interval is designable. As a result, the model is trained on both fragment-completion and fragment-
  replacement style conditioning patterns. Each sampled training example therefore contains both fixed and designable atoms, and the mask is incorporated directly into the model input and loss masking.

  As for the unconditional task, the metrics for the completion task evaluate the full molecule distribution (i.e. not just the completed part), which should be learned if the completion task is correctly learned.

\begin{figure}[t]
    \centering{\includegraphics[width=0.99\textwidth]{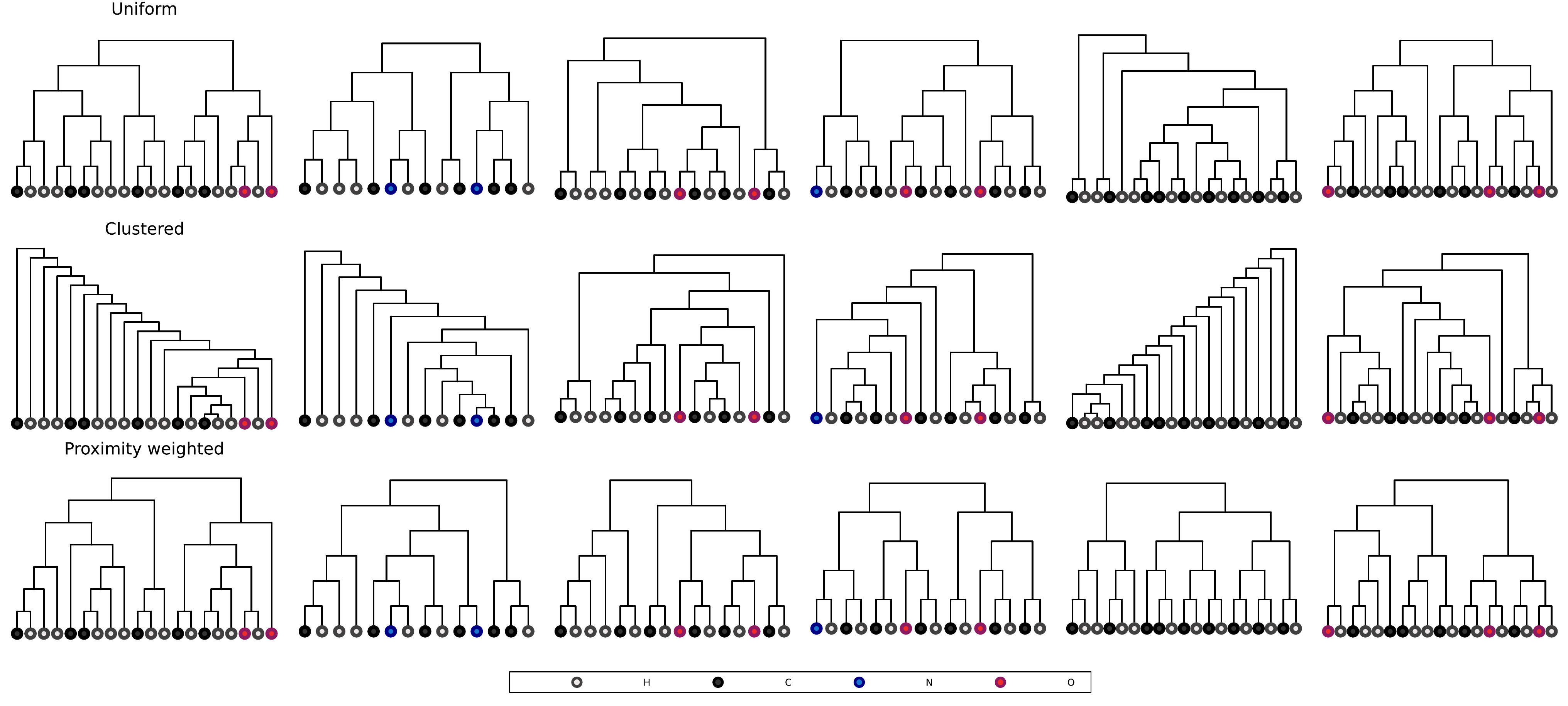}}
    \caption{\textbf{Trees policies.} Shown are the default `uniform adjacent' tree sampling, vs the Rich-get-richer (here labelled as `Clustered') and Proximity Weighted strategies, described in Appendix \ref{SI:variations}.}
    \label{fig:treepolicies}
\end{figure}

\begin{figure}[b!]
    \centering{\includegraphics[width=0.99\textwidth]{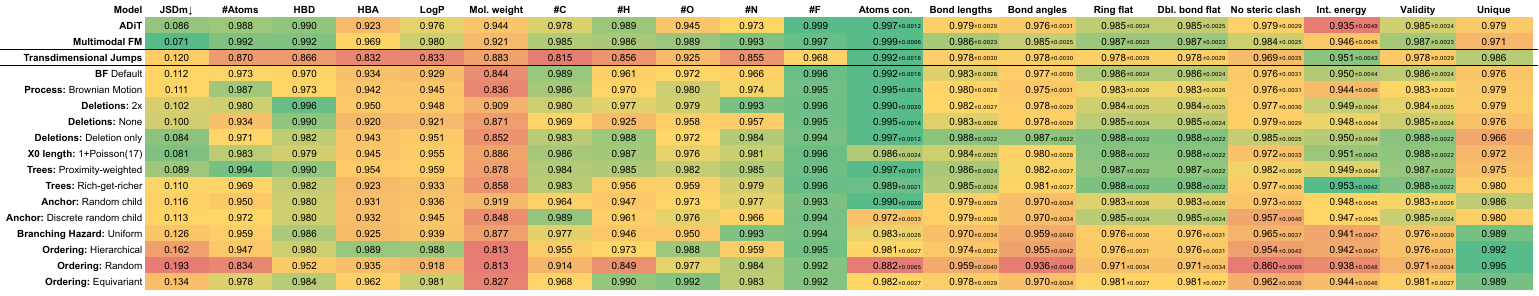}}
    \caption{\textbf{QM9: Variations and Ablations.} Description of model variants is in \ref{SI:variations} and metrics are the same as from Figures \ref{fig:QM9maintable} and \ref{fig:posebusters}. All variations and ablations here (excluding ADiT and Transdimensional Jumps) were trained for 400 iterations.}
    \label{fig:variations}
\end{figure}

\subsection{Branching Flows QM9 variations and ablations}
\label{SI:variations}

For the results in figures \ref{fig:QM9maintable} the Branching Flows and architecture matched comparison models were trained for 800k iterations for the unconditonal task and 400k iterations for the conditonal tasks. For the full set of ablations and variations (described below, and shown in Appendix figure \ref{fig:variations}) the models were trained for 400k iterations.

Here we describe all model variants. Unless a change is stated explicitly in a variation subsection below, the setting is identical to the Default model and process described above:

\paragraph{\texttt{Default}.}
As described above.

\paragraph{\texttt{Multimodal FM}.}
This is the fixed-length comparison. No splits or deletions, but the process, training, and architecture otherwise matches the `Default' BF model. This model always knows the `correct' molecule length.

\paragraph{\texttt{Brownian Motion}.}
The Ornstein--Uhlenbeck process is replaced by Brownian motion (variance = 1).

\paragraph{\texttt{Deletions: 2x}.}
The deletion padding is increased from 20\% to 100\% (ie. on average the $X_1$ length is doubled), so the bridge is trained with substantially more to-be-deleted leaves.

\paragraph{\texttt{Deletions: None}.}
Deletion is disabled by setting \texttt{deletion\_pad}$=0$ and removing the deletion head from both training and sampling. This learns length variation via splits only.

\paragraph{\texttt{Deletions only}.}
This ablation removes splits entirely, and begins with an $X_0$ length distribution that is Uniform between 30 and 50 (i.e. guaranteed to always be longer than the $X_1$ length, which is required for distribution learning with only deletions). We note that, since this model is computing over sequences that are typically longer the standard Branching Flows models, the training time (quadratic in length for transformers) is slower.

\paragraph{\texttt{X0 length: 1+Poisson(17)}.}
The initial length prior is changed to $1+\mathrm{Poisson}(17)$, no longer starting from a single element (and approximately matching the target distribution length mean).

\paragraph{\texttt{Trees: Proximity-Weighted}.}
The Default variant randomly merges adjacent elements when constructing trees. The Proximity Weighted variant still uses adjacent pairs, but instead of sampling them uniformly it computes a distance between the extracted node features and chooses the adjacent pair with minimum distance, with random tie-breaking inside a small tolerance band. Merges are biased toward geometrically nearby adjacent atoms. When merging internal nodes it uses the anchors, which are weighted averages of child nodes.

\paragraph{\texttt{Trees: Rich-get-Richer}.}
This still considers only eligible adjacent pairs, but it samples pair $(i,i+1)$ with probability proportional to
$(w_i + w_{i+1})^\alpha$
where $w_i$ and $w_{i+1}$ are the descendant counts of the two candidate subtrees. With $\alpha=1$, larger adjacent clusters are therefore more likely to merge first.

\paragraph{\texttt{Anchors: Random Child}.}
Instead of a weighted average for continuous states, and a mask token for discrete states, for continuous states this chooses one of the two child anchors with probability proportional to the child descendant counts and then copies that child exactly for continuous components instead of averaging the two children. Discrete anchors remain masked.

\paragraph{\texttt{Anchors: Discrete random Child}.}
Continuous anchors use the canonical weighted average, but discrete anchors copy one child atom-type state with probability proportional to child weight instead of being replaced by the masked token.

\paragraph{\texttt{Branching Hazard: Uniform}.}
The branch-time distribution is changed from $\mathrm{Beta}(1,3/2)$ to $\mathrm{Uniform}(0,1)$.

\paragraph{\texttt{Ordering: Heirarchical}.}
To test whether the canonical SMILES ordering is critical, the default order is removed. For each sampled molecule, the atoms are first randomly permuted, then clustered by Euclidean average-linkage hierarchical clustering, and finally linearized by recursively swapping the left and right children of each internal node at random. This yields a sequence order that respects the hierarchical geometric clustering while still changing stochastically each time the molecule is used. The Proximity Weighted tree policy is used.

\paragraph{\texttt{Ordering: random}.}
The default order is removed. Instead, every time a molecule is sampled for training, the atom order is replaced by a fresh random permutation. The Proximity Weighted tree policy is used.

\paragraph{\texttt{Ordering: Permutation Equivariant}.}
This variation removes explicit sequence-order dependence from both the neural network and the tree policy. RoPE is removed from the transformer. The tree construction policy is examines all eligible intragroup pairs $(i,j)$ rather than only adjacent pairs and selects the pair with minimum Euclidean distance between the extracted node coordinates, again with random tie-breaking among near-ties.

\subsubsection{QM9 compute resources}
Each model training run of 400k iterations takes approximately one day on an RTX 6000 Ada, and two days for the 800k training runs.

\begin{figure*}
    \centering{\includegraphics[width=0.99\textwidth]{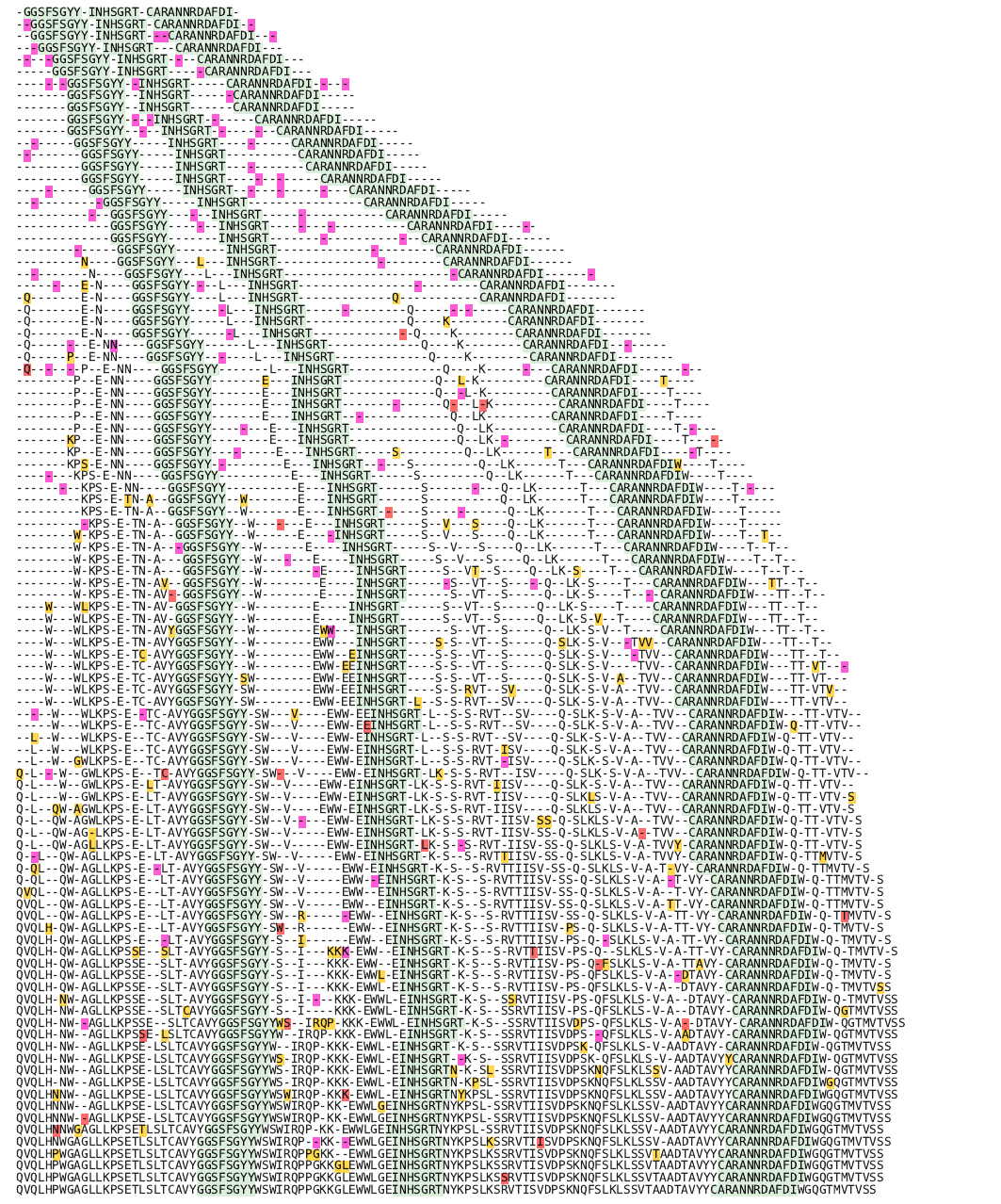}}
    \caption{\textbf{Antibody segment sampling.} A sample from the Branching Flows model trained for the conditional antibody task, where multiple random-length segments were fixed, and the model generates the non-fixed segments. For this example, we fixed the complementarity determining regions and the model successfully samples four framework regions, including two outer flanking regions and two infix regions. Green shows fixed regions. Magenta shows insertions, yellow shows substitutions, and red shows elements that are deleted in the next step.}
    \label{fig:condabexample}
\end{figure*}

\subsection{Antibodies}\label{dat:abs}

\subsubsection{Data}
Antibody sequence models were trained on a filtered subset \cite{10.1093/bioinformatics/btae659} of the antibody heavy chains from the Observed Antibody Space database (OAS) \citep{kovaltsuk2018oas} (license: CC-BY-4.0). The filtering retained only human sequences and removed sequences with unknown residues, missing conserved cysteines and those which had framework region 1 or 4 shorter than allowed by IMGT definitions. Sequences were further filtered to remove redundancy by clustering at 95\% identity. This resulted in a total of 118 million sequences, of which a random subset of 15 million were sampled and used for training. For the evaluation of the model, a similarly-filtered subset of the OAS data was used. 

\begin{figure*}[t]
    \centering{\includegraphics[width=0.99\textwidth]{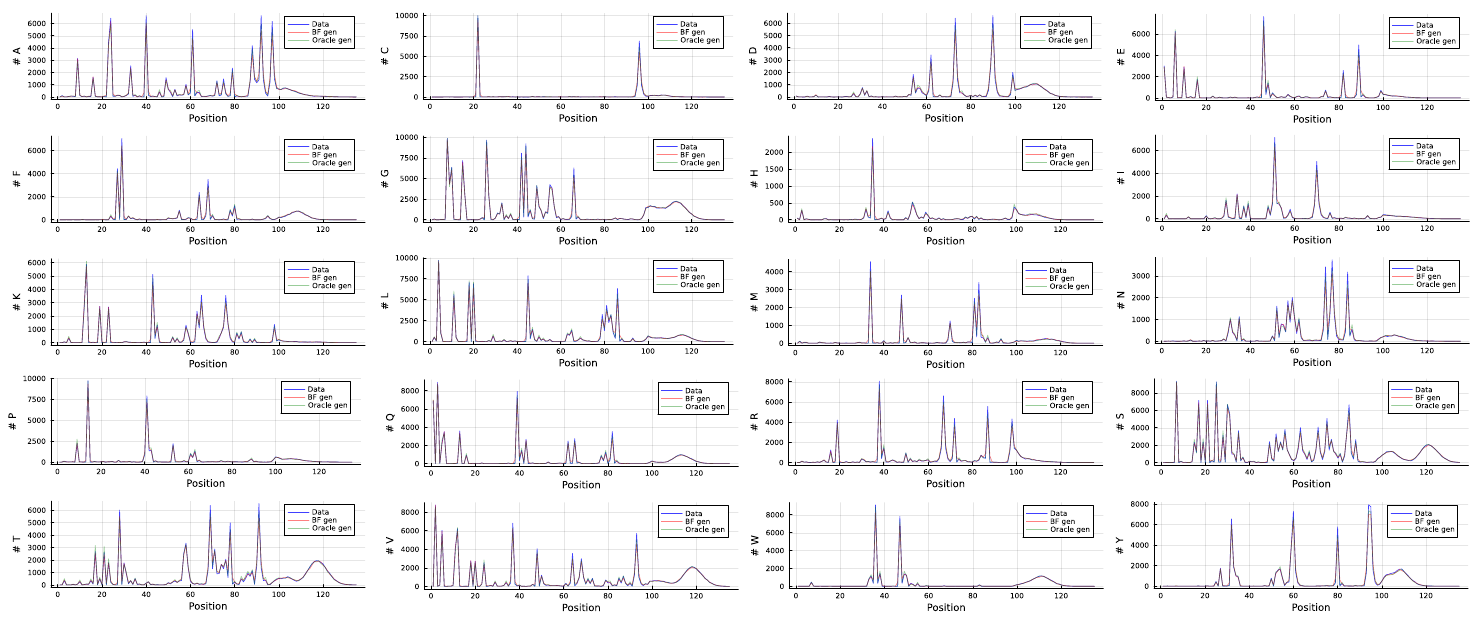}}
    \caption{\textbf{Amino acid position distributions.} The count of the number of each amino acid at every position in natural data, and in samples from Branching Flows (`BF gen') and Oracle Length DFM (`Oracle gen').}
    \label{fig:abpos}
\end{figure*}

\subsubsection{Branching Flows Specification}
\begin{outline}
 \1 Conditional base process:
        \2 Discrete: DFM convex interpolation (appendix \ref{app:dfm}) of $x_0$, uniform noise, and $x_1$, with $\omega_u = 0.2$, and CDFs $F_1$ and $F_2$ of $\mathcal{D}_1=\text{Beta}(1.5,1.5),$ and $\mathcal{D}_2=\text{Beta}(2,2),$ respectively.  
\1 Branching hazard distribution: $\operatorname{Beta}(1,2)$
\1 Deletion hazard distribution: $U(0,1)$
\1 To-be-deleted policy:  deletion rate $d_r = 1.2$
\1 $x_0$ state distribution: masked tokens.
\1 $x_0$ length distribution: uniform between 110 and 140.
\end{outline}

\subsubsection{Model and Training}
For the antibody sequence modeling task, we used a transformer-based architecture similar to that of a diffusion transformer \citep{peebles2023sdmt}, but with additional split and deletion heads. Time was embedded with RFFs, and adaptive layer normalization (adaLN-Zero) was used for time conditioning. The model had an embedding dimension of 768, 12 attention heads, and 8 layers.

The models (Branching Flows, oracle-length and Edit Flows) were trained using a batch size of 64 resulting in a total of $234\,375$ training steps for the 15 million sequence dataset. Muon was used with a 2000 step linear warm-up followed by a cosine decay starting at $10^{-3}$ and decaying to $10^{-7}$. 

\subsubsection{Antibody Sequence Generation: Metrics and Comparisons}
To evaluate Branching Flows, $10\,000$ sequences were generated with $1\,000$ uniformly-spaced steps from Branching Flows, Oracle Length, and Edit Flows models and then compared to natural ones. Sequence diversity was assessed by computing the minimum pairwise cosine distance (over vectors of 3-mer counts) within each set. Sequence `novelty' was evaluated by computing the minimum Levenshtein distance to any sequence in a hold-out validation data set (needed because otherwise the novelty of all natural sequences would be zero).

Sequence `nativeness' was computed using AbNatiV \cite{ramon2024abnativ}. Finally, the distribution of amino acids, distribution of sequence lengths and the distribution of CDR3 lengths were also computed. The latter was computed using ANARCI \cite{dunbar2016anarci}.

For investigating the training dynamics of Branching Flows compared to Oracle Length and Edit Flows models the losses cannot be directly compared, so after every 500 steps during training, 5 sequences were generated (with $200$ uniformly-spaced steps) and saved. The perplexity of the generated sequences was computed as evaluated by a 130 million parameter autoregressive model trained on the data from \cite{10.1093/bioinformatics/btae659}.

\begin{figure}[t]
    \centering{\includegraphics[width=0.99\textwidth]{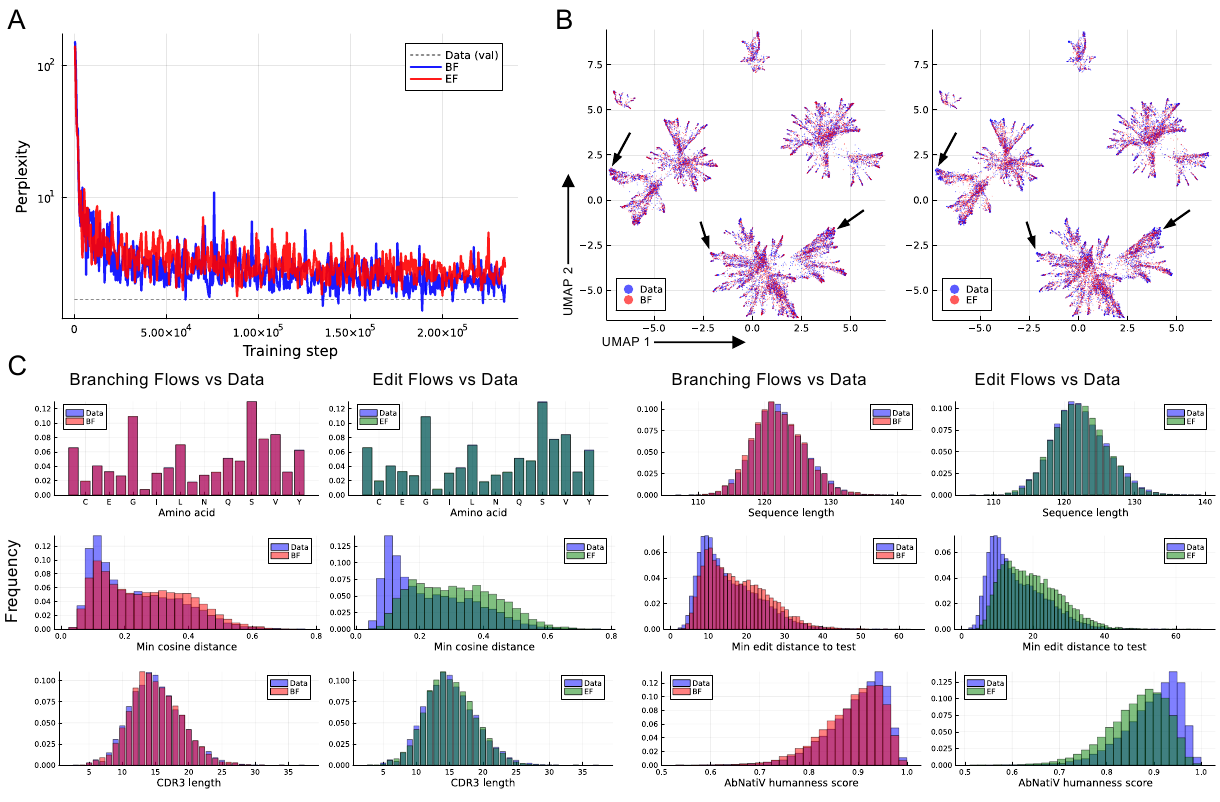}}
    \caption{\textbf{Branching Flows vs Edit Flows.} 
    \textbf{A} The perplexity, evaluated under an autoregressive LLM, of samples generated by Branching Flows (BF) vs Edit Flows (EF), which is a discrete-only variable length flow matching strategy. This is shown over training iterations. \textbf{B} two seqUMAP \citep{hanke2022multivariate} plots comparing Branching Flows and Edit Flows (respectively) each against real sequences. Arrows point to regions, at the extremities of the UMAP-embedded shapes, where Edit Flows density appears to be reduced. \textbf{C} Comparison of several distributions of the generated sequences with sequences from the validation data set. 
    }
    \label{fig:bf_vs_ef}
\end{figure}

For the antibody conditional `completion' task, for each heavy-chain sequence, a set of contiguous editable regions was sampled, while the remainder of the sequence was treated as fixed context, meaning that insertions, deletions, and substitutions were disallowed outside the editable regions. The number of editable regions was sampled as $1 + \mathrm{Poisson}(6 \cdot U(0,1))$ and the length of each proposed region was sampled as $\mathrm{Poisson}(30 \cdot U(0,1))$. For each proposal, a starting position and a direction were sampled uniformly, and the final editable region was the union of all spans. Overlapping proposals therefore merged into contiguous editable regions.
 
For each contiguous editable region, in contrast to the unconditional Branching Flows setup, the completion task does not use a global $x_{0}$ length prior, but instead initialized each editable region from exactly one token (for both Branching Flows and Edit Flows). Editible regions were augmented with mean 30\% `to-be-deleted' states for Branching Flows.
 
The same 15 million antibody sequences were used as in the unconditional experiments. The model scale was preserved (hidden dimension 768, 12 attention heads, 8 layers), and the only difference in the models relative to the prior task was that the conditional models were conditioned on binary masks representing which positions were editable. Training hyperparameters were also matched to the unconditional setting: a batch size of 64 was used with a 2000-step linear warm-up to $10^{-3}$ and then a cosine decay to $10^{-7}$. Finally, the models were evaluated by generating 10 000 sequences using the same conditional specification used during training. The fixed contexts were sampled from a held-out test set. Moreover, to quantify how well the sequences aligned with true sequences, the same distribution matching metrics as used for the unconditional task were computed using the entire sequences (not just the editable regions).

\subsubsection{Antibody compute resources}
Each antibody model training run takes approximately 3.5 days on a single Nvidia A6000 GPU.

\subsection{Proteins}
\label{dat:prot}

To investigate whether Branching Flows would scale to a more complex learning problem where the state space is multimodal, we modified and finetuned a pre-trained ChainStorm model \citep{OrestenChainStorm} to allow the model to determine how long chains, or designable segments, should be.

Briefly, ChainStorm is a multimodal flow matching model that jointly generates protein backbones and amino acid sequence labels (but not side chain atom positions). Amino acid backbones are represented as sequences of `frames' \citep{jumper2021highly} with Euclidean positions and SO(3) rotations, and ChainStorm uses an Invariant Point Attention \citep{jumper2021highly} transformer architecture to estimate the $t=1$ frame positions and orientations, whose conditional/marginal paths are governed by Euclidean Brownian Motion and SO(3) rotational diffusion. Discrete Flow Matching \citep{Gat2024DiscreteFM} is used for amino acid sequence labels.

ChainStorm's Euclidean and SO(3) conditional processes were constructed with relatively low noise, but we reasoned that Branching Flows, especially with late-time branching events requiring higher flow velocities, might benefit from noisier processes which would have higher flow velocities generally, so we first finetuned the default ChainStorm model (weight license: MIT) with modified element-wise conditional processes, using
\begin{itemize}
\item for Euclidean positions, a time-inhomogeneous OU process (appendix \ref{app:time-inhomo-ou}) with mean reversion rate of 100, and a variance of 150 at $t=0$ that decays to $0$ as $t \rightarrow 1$. 
\item for SO(3) rotations, a manifold-respecting version of the above OU process with identical parameters.
\item for discrete states, a DFM convex interpolation (appendix \ref{app:dfm}) with a $\mathcal{D}_1=\text{Beta}(3.0,1.5),\mathcal{D}_2=\text{Beta}(2,2), \omega_u=0.2$.
\end{itemize}
All Branching Flows protein models were then finetuned atop this model, which exhibited much higher variance sampling paths than the original ChainStorm.

We then modified the ChainStorm architecture to construct `BF-ChainStorm' to allow for i) element branching and deletion, and ii) conditioning on fixed protein components (in general any set of residues, but here chains or contiguous segments of chains) that are then prevented from changing during training or inference. Conditioning is done by including a `conditioning mask' embedding layer whose outputs get summed into the initial activations, and by preventing the per-layer frame updates for fixed residues. For element branching and deletions, we reasoned that, since we are finetuning into an existing model, we didn't want to only extract contributions from the final output layer (since this is used for rotations, positions, and discrete states), so we concatenate the output activations from the last three layers. This is summed into an embedding of $t$ and that sum is passed into a SwiGLU-like layer that collapses to a single dimension. This is done separately for branching and deletion, and the outputs are interpreted as `logits' for loss (during training) and rates (during inference). Finally, since the state dimension now changes during inference, we replaced ChainStorm's self-conditioning with recycling, which requires no architectural changes but instead requires additional model passes within each inference step.

One subtlety related to Branching Flows is that, originally, ChainStorm conditioned on residue indices which provide, during training, the model with information about when parts of backbone are missing (i.e. unresolved in the crystal or cryo-EM structures) since a gap in the residue numbering will be present. To generate backbones without `chain breaks' (usually the desired behavior), at inference time the generation is conditioned on residue numbers with no gaps. However, since BF-ChainStorm's conditional/marginal paths change by insertions and deletions with $t$, including residue numbers no longer makes sense, as the number of residues for any intervening segment (which, in BF-ChainStorm, must be stochastically generated during inference) would be pre-determined by the residue numbering. So instead of residue indices, we use fixed-increment positional index. However, since the training data has many chain breaks, the model would then stochastically generate structures with chain breaks. So we additionally introduced a `break mask' encoder (also summed into the starting activations) that encodes, on a per-chain basis, whether the `designable' portion of the chain includes any chain breaks. When used during inference, this reduces the number of chain breaks but doesn't remove them entirely, possibly because the model is getting conflicting signals between the conditioner and the emerging structure (e.g. large unstructured loops are often not resolved in protein structures, so if the model is generating one that might overwhelm the chain-break conditioner). Chain breaks are easy to detect using bond distances, and we exclude them from refolding analysis because our folding pipeline requires intact chains.

When sampling trees $T$ for protein models, when the group (described above) is the chain index, the requirement that separate chains never occur on the same tree allows the number of sampled chains in a structure to be specified at inference time (but each of their lengths stochastically sampled by the model). Similarly, any residues (even within a single chain) separated by residues fixed in the conditioning (relevant for BranchSegment) may not coalesce when sampling trees. Thus no single tree spans two designable segments.

For BranchChain, we initialize each chain to have $1+\text{Poisson}(20)$ $x_0$ elements, each drawn from the ChainStorm base element distribution, and for BranchSegment each designable segment has a single $x_0$ element. For sampling `to-be-deleted` elements, we use $d_r=1.2$ for BranchChain and $dr=1.1$ for BranchSegment (reasoning that the more strongly informative conditioning information for BranchSegment might require less exploratory flow paths). Samples were generated with 400 steps: $t = 1-(\cos(\pi\tau)+1)/2$ for $\tau \in (0, 1/400, 2/400, ..., 1).$

For illustration purposes, we also finetuned a BranchChain variant with only a single $x_0$ element per chain. Figure \ref{fig:prottraj} shows steps along the sampling trajectory for a sample where two chains (blue and green) were fixed, and two chains were sampled. The orange chain, best seen through the lens of the model-predicted end state $x_1^{hat}$, highlights a key Branching Flows feature: at first, the designed chain converges upon two separate domains, with insufficient residues in between them for them to be linked. This is then progressively filled in by insertions, until a complete linker connects them as a single chain.

For BranchSegment we do not attempt quantitative evaluations, but we demonstrate that it is a promising solution to the `unknown length infix sampling' problem. Taking a protein complex (PDB: 9IQP, \citet{shcheblyakov2025ultra}) of a nanobody and the SARS-CoV-2 receptor binding domain (RBD), we mask only the `Complementarity Determining Region 3' (CDR3) as designable, and we generate 20 samples from the RBD-bound complex, and 20 samples from the unbound nanobody alone. The sampled CDR3s had a broad length distribution (7 to 29 residues for bound, 7 to 40 unbound). For the bound samples, even though the side chain atom positions from the RBD were not provided to BranchSegment (it only sees backbone frames and amino acid labels) the designed CDR3s had no atomic clashes with the RBD side chain rotamers for 15 out of 20 of the generated CDR3s. Some of the unbound CDR3s appear structurally plausible, but are extremely unlikely to occur naturally, which is not surprising since neither the base model nor BranchSegment have any antibody or nanobody-specific training.

\subsubsection{Branching ChainStorm compute resources}
Each ChainStorm finetune takes approximately 3 days on a single Nvidia RTX 6000 Ada GPU.

\section{Branching La Proteina}

\subsection{La Proteina Branching Flows fine-tuning and evaluation}

  \subsubsection{Branching-flow architecture}
  We fine-tuned the unconditional La Proteina LD3 model (weights license: NVIDIA Open Model License Agreement) with Branching Flows. The base network was the 160M-parameter local-latents transformer with 14 transformer layers, token dimension 768, 12
  attention heads, conditioning dimension 256, and output parameterization as vector fields for both C$_\alpha$ coordinates and local latent variables. The Branching Flows variant wrapped this base transformer and added only the machinery required to model insertions and deletions.

  Two additional per-residue heads were attached to the final sequence embeddings of the LD3 trunk: a split head and a deletion head. Each indel head was a two-layer MLP,
  \[
  768 \rightarrow 768 \rightarrow 1,
  \]
  with a SiLU nonlinearity between layers. The first linear layer used a bias and the final scalar projection did not.
  The C$_\alpha$ and local-latent vector-field output heads were otherwise the original LD3 heads. Unlike the original La Proteina, the Branching Flows variant used the same scalar time \(t\) for C$_\alpha$ and local-latent conditioning during Branching Flows training and sampling.
  
  \subsubsection{OU bridge process and Branching Flows setup}
  Training bridges and Branching Flows sampling used an endpoint-conditioned Ornstein--Uhlenbeck process (as above) rather than the ODE process from the original model. Following the insight from La Proteina \citep{geffner2025laproteinaatomisticproteingeneration} that co-designability improves when the latent side chains are resolved later in the conditional process, we chose OU parameters to give similar behavior.
  For C$_\alpha$ coordinates during training, we used the same OU process as ChainStorm, but for the latent side chains we used a mean reversion rate of 5, and a constant variance of 5 (i.e. we do not decay the variance, and the termination at the endpoint is due to the conditioning drift only).

  The branching-time hazard distribution was a \(\mathrm{Beta}(1,2)\) distribution rescaled to support \([0,0.95]\) to prevent late split events (which was identified as an issue specific to La Proteina in an initial training attempt). To-be-deleted elements were sampled from a Poisson with mean 10\% of the length of the training sample.

  \subsubsection{Fine-tuning setup}
  The model was initialized from the LD3 checkpoint. The (frozen) VAE from LD3 was usedto generate side chain latents for the Branching Flows finetune. 

  Fine-tuning was run for 1,000,000 training steps on one GPU with batch size 8 (via gradient accumulation). Training used full float32 precision. Self-conditioning was enabled: before the gradient pass, a Poisson(\(1\)) number of no-gradient self-conditioning passes was sampled, and the resulting predicted endpoints were fed back as self-conditioning inputs.

  The optimization used Muon for the transformer stack matrix parameters and AdamW for the remaining parameters, including the input/conditioning factories, output heads, and the new indel heads. Muon used momentum 0.95, Nesterov momentum,
  weight decay 0.01, \(\epsilon=10^{-7}\), and 5 Newton--Schulz steps. AdamW used learning rate \(0.015\) times the Muon learning rate, betas \((0.9,0.95)\), \(\epsilon=10^{-10}\), and weight decay 0.01. Gradients were clipped to norm 1.0.

  The learning-rate schedule started at \(5\times10^{-6}\), inflated by a factor 1.05 at each learning-rate update until reaching the target \(10^{-4}\), and then decayed multiplicatively by 0.99995. Learning-rate updates occurred every 10 optimizer updates. A final warmdown was enabled for the last 7,810 optimizer updates, linearly decaying the learning rate to \(10^{-9}\).

  \subsubsection{Fixed-length La Proteina LD3 baselines}
  For comparison, we generated fixed-length samples from the original LD3 model using lengths matched one-to-one to the 500 Branching Flows samples. Each fixed-length LD3 sample was indexed by the corresponding Branching Flows sample it matched.
  La Proteina, like most other protein models, uses annealed noise schedules to improve designability when sampling.
  Two LD3 noise settings were evaluated: the default La Proteina inference setting from the codebase \((0.15,0.05)\) for C$_\alpha$ and local latents, and an alternative \((0.1,0.1)\) which is what was used in the La Proteina manuscript.

  The original fixed-length LD3 sampler used 400 steps and self-conditioning. The C$_\alpha$ and local-latent modalities used different time grids. C$_\alpha$ used the log schedule
  \[
  t_i = 1 - \mathrm{logspace}(-2,0,N+1)^{\mathrm{reversed}},
  \]
  renormalized to span \([0,1]\). Local latents used a power schedule \(t_i=(i/N)^2\). C$_\alpha$ used \(g(t)=1/(t+10^{-2})\), while local latents used
  \[
  g(t)=\frac{\pi}{2}\frac{\sin((1-t)\pi/2)}{\cos((1-t)\pi/2)+10^{-2}}.
  \]
  Both modalities used stochastic score-corrected sampling mode. For \(t\leq0.98\), the update was
  \[
  x_{t+\Delta t}=x_t+\{v_\theta(x_t,t)+g(t)s_\theta(x_t,t)\}\Delta t
  +\sqrt{2g(t)\lambda\Delta t}\,\epsilon,
  \]
  where \(\lambda\) was the modality-specific noise scale. For \(t>0.98\), the implementation switched off stochastic noise and used a low-temperature ODE update with the inferred score scaled by 1.5. C$_\alpha$ coordinates were re-centered
  after every step; local latents were not.

  The main text table shows metrics for samples from the \((0.15,0.05)\) model (which performed better on our metrics) and table \ref{laprot_SItable} shows both LD3 annealing options.

\subsubsection{Branching-flow sampling and annealing}
  For the final Branching Flows LD3 samples reported here, we used the fine-tuned checkpoint above with 500 sampling steps, self-conditioning enabled. Sampling used the cosine time grid as in the Branching Flows ChainStorm finetune.
  Annealing for Branching Flows models is underexplored, and here we modified our OU noise (just as the original La Proteina modifies their SDE noise) to achieve a similar effect.
  The C$_\alpha$ OU process used a flat annealing value \(v_0=60\), with \(v_1=10^{-9}\) held near zero. The local-latent OU process (calibrated to resolve later) used \(\theta=5\), \(\beta=-0.001\), and linearly annealed variance from 5 at \(t=0\) to 0 at \(t=1\).

  \subsubsection{Designability, co-designability, and diversity metrics}
  All reported metric sets used \(N=500\) generated structures. MPNN1 designability was computed by running ProteinMPNN once per generated backbone, folding the single designed sequence with ESMFold, and computing C$_\alpha$ TM-score between the
  ESMFold structure and the generated structure. A sample was counted as designable when \(\mathrm{scTM}>0.5\) \citep{lin2023generatingnoveldesignablediverse}

  Co-designability used the original sequence sampled from the generated latent side chains rather than a ProteinMPNN-designed sequence. This sequence was folded with ESMFold and compared back to the generated structure using the same C$_\alpha$ TM-
  score criterion. A sample was counted as co-designable when \(\mathrm{scTM}>0.5\).

  Diversity was computed with Foldseek \citep{vanKempen2024Foldseek} clustering on the generated PDBs. The reported structure-only diversity used Foldseek \texttt{easy-cluster} with \texttt{--alignment-type 1}, \texttt{--cov-mode 0}, \texttt{--min-seq-id 0.0}, and
  \texttt{--tmscore-threshold 0.5}. The reported structure-plus-sequence diversity used \texttt{--alignment-type 2}, \texttt{--cov-mode 0}, \texttt{--min-seq-id 0.1}, and \texttt{--tmscore-threshold 0.5}. For binomial designability and co-
  designability proportions, the table reports symmetric two-sigma Wald intervals.

\begin{table}[h]
  \centering
  \begin{tabular}{lcccc}
  \hline
  Sample set & MPNN1 designability & Co-designability & Div. str & Div. str+seq \\
  \hline
  Branching Flows
  & $0.990 \pm 0.009$
  & $0.966 \pm 0.016$
  & 128 & 215 \\
  LD3 matched, $0.15,0.05$
  & $0.972 \pm 0.015$
  & $0.952 \pm 0.019$
  & 110 & 173 \\
  LD3 matched, $0.1,0.1$
  & $0.964 \pm 0.017$
  & $0.950 \pm 0.019$
  & 96 & 147 \\
  \hline
  \end{tabular}
  \caption{Designability, co-designability, and diversity for $N=500$ samples. Intervals are symmetric two-sigma Wald intervals for binomial proportions.}
  \label{laprot_SItable}
\end{table}

\subsubsection{Branching La Proteina compute resources}

Finetuning Branching Flows into the La Proteina LD3 model takes approximately one day on an Nvidia RTX 6000 Blackwell.

\endgroup
